\RequirePackage{fix-cm}
\documentclass[smallextended]{svjour3}       
\smartqed  
\usepackage{bm}
\usepackage{float}
\usepackage{amsmath}
\usepackage{graphicx}
\usepackage{multirow}
\usepackage{subfigure}
\usepackage[table]{xcolor}
\usepackage[colorlinks, citecolor=green]{hyperref}
\usepackage[ruled,norelsize]{algorithm2e}

\begin{document}

\title{A Federated Data-Driven Evolutionary Algorithm for Expensive Multi/Many-objective Optimization}

\thanks{Grants or other notes
about the article that should go on the front page should be
placed here. General acknowledgments should be placed at the end of the article.}



\author{Jinjin Xu         \and
        Yaochu Jin          \and
        Wenli Du
}


\institute{Jinjin Xu \at
          East China University of Science and Technology, Shanghai, China. \\
          \email{jin.xu@mail.ecust.edu.cn}             \\
          \emph{Present address: University of Surrey, Guildford, UK.} of Jinjin Xu  
          \and
          Yaochu Jin \at
          University of Surrey, Guildford, UK.
          \and
          Wenli Du \at
          East China University of Science and 
}

\date{Received: date / Accepted: date}

\maketitle

\begin{abstract}
Data-driven optimization has found many successful applications in the real world and received increased attention in the field of evolutionary optimization. Most existing algorithms assume that the data used for optimization is always available on a central server for construction of surrogates. This assumption, however, may fail to hold when the data must be collected in a distributed way and is subject to privacy restrictions. This paper aims to propose a federated data-driven evolutionary multi-/many-objective optimization algorithm. To this end, we leverage federated learning for surrogate construction so that multiple clients collaboratively train a radial-basis-function-network as the global surrogate. Then a new federated acquisition function is proposed for the central server to approximate the objective values using the global surrogate and estimate the uncertainty level of the approximated objective values based on the local models. The performance of the proposed algorithm is verified on a series of multi/many-objective benchmark problems by comparing it with two state-of-the-art surrogate-assisted multi-objective evolutionary algorithms. 
\keywords{Federated learning \and Multi/Many-objective optimization \and Surrogate \and Data-driven evolutionary optimization \and Bayesian optimization}
\end{abstract}

\section{Introduction}
\label{intro}
Many optimization problems have several conflicting objectives that need to be optimized concurrently, which are known as multi-objective optimization \cite{coello2020multi}, and many-objective optimization when the number of objectives is larger than three \cite{li2015many}. Over the past decades, a pile of successful multi-objective evolutionary algorithms (MOEAs) have been developed, which can largely be categorized into decomposition-based MOEAs, such as the multi-objective evolutionary algorithm based on decomposition (MOEA/D) \cite{zhang2007moea} and reference vector guided evolutionary algorithm (RVEA) \cite{cheng2016reference}; Pareto-dominance based, e.g., the elitist non-dominated sorting genetic algorithm (NSGA-II) \cite{deb2002fast}; and performance indicator-based \cite{emmerich2005sms}. Note that many MOEAs combine decomposition or Pareto dominance with other criteria to address various challenges in multi-objective optimization \cite{zhang2015knea}.  
   
It is well recognized, however, that most MOEAs require a large number of function evaluations before a set of diverse and well-converged non-dominated solutions can be found, which makes it hard for them to be directly applied to solve a class of data-driven optimization problems \cite{jin2018data}, whose objectives can be evaluated by means of conducting time-consuming computer simulations or costly physical experiments. Examples of real-world data-driven multi-objective optimization problems include blast furnace optimization \cite{chugh2017blast}, air intake ventilation system optimization \cite{chugh2017intake}, airfoil design optimization \cite{li2020data}, bridge design optimization \cite{montoya2018cfd}, design optimization for a stiffened cylindrical shell with variable ribs \cite{zhou2019two}, optimization of operation of crude oil
distillation units \cite{guo2021evolutionary}, resource allocation of trauma systems \cite{wang2016trauma}, and also neural architecture search \cite{sun2020cnn,zhou2020lsm}, among many others.  


Data-driven surrogate-assisted evolutionary algorithms were first developed for single-objective optimization \cite{jin2005comprehensive}. In recent years, more attention has been paid to data-driven surrogate-assisted multi-objective optimization \cite{allmendinger2017survey,chugh2019survey}. Similar to MOEAs, surrogate-assisted MOEAs can also be divided into four groups. The first group is based on decomposition, where surrogates are built to approximate the sub-problems using random weights, known as ParEGO \cite{knowles2006}. Alternatively, surrogates can also be constructed for each individual objective function to assist a decomposition based MOEA, such as the Gaussian process assisted MOEA/D, called MOEA/D-EGO \cite{zhang2009expensive}, K-RVEA \cite{chugh2016surrogate}, a Kriging-assisted RVEA, and surrogate-ensemble assisted MOEA, called SAEMO \cite{zhao2021surrogate}. The second group employs the non-dominance relation for environmental selection, and surrogates are adopted for approximating the objective functions \cite{wang2020forest,guo2018heterogeneous}, or in local search \cite{lim2010}. In \cite{loshchilov2010,pan2019class}, a surrogate is built for estimating the dominance relationship, and consequently only a single surrogate is required. The third category relies on a performance indicator-based MOEA, e.g., a Gaussian process assisted hypervolume based MOEA \cite{emmerich2006}, and an enhanced inverted generational distance based MOEA assisted by a dropout deep neural network (EDN-ARMOEA) \cite{guo2021evolutionary}. Finally, surrogates have also been applied to assist other MOEAs, such as the surrogate-assisted two-archive evolutionary algorithm \cite{wang2021two}. A good overview of data-driven evolutionary optimization can be found in a recently published book \cite{jin2021book}.

Despite its impressive success, data-driven evolutionary optimization is faced with several challenges. One is the lack of data, which is very expensive to collect. To address this issue, advanced learning techniques such as semi-supervised learning \cite{sun2013semi,sun2021ppsn,wang2021tri}, knowledge transfer between different surrogates using transfer optimization \cite{yang2019,wang2020applied}, and between different objectives using transfer learning \cite{wang2020gecco,wang2021kbs} have been adopted in surrogate-assisted optimization. Another solution is to reduce the computational cost of data collection by parallelizing expensive computer simulations \cite{akhtar2019efficient,briffoteaux2020} in data-driven surrogate-assisted evolutionary optimization. 

Little attention has been paid to data privacy concerns in data-driven optimization. One exception is our previous work reported in \cite{xu2021federated}, which proposes a federated data-driven evolutionary algorithm (FDD-EA) for single-objective optimization based on federated learning \cite{mcmahan2017communication,zhu2021federated}. As its name suggests, federated learning aims to collaboratively train a global model using data distributed on multiple local devices without transmitting the local data to a central server. Instead, each client train a local model based on the local data and then the parameters (or the gradients) of the local models are transmitted to the server to generate the global model. Similarly, in FDD-EA, the central server averages the local models uploaded by the clients with a sorted averaging algorithm  \cite{xu2021federated}. Then, the global and local models are made use of for approximating the objective function and estimating the uncertainty of the approximated fitness, based on which an infill criterion is designed for surrogate management in the federated optimization environment. By extending FDD-EA, this work proposes a federated data-driven evolutionary algorithm for multi-objective optimization. The main contributions of this paper are summarized as follows.

\begin{itemize}
    \item A federated data-driven multi-/many-objective evolutionary optimization algorithm (FDD-MOEA) is presented, in which multiple clients collaboratively train a global surrogate without uploading local data. To the best of our knowledge, this is the first work that enables federated data-driven multi/many-objective optimization that does not require to store the local data on a central server, thereby preserving the privacy of the data.
    
    \item An ensemble-based acquisition function is adopted for FDD-MOEA, in which the global surrogate is used to provide the approximated fitness and the local surrogates are used to provide an estimate of the uncertainty of the approximated fitness. 

\end{itemize}

The remainder of this paper is arranged as follows. Section \ref{sec:related} presents the concepts of multi-objective optimization, federated learning and acquisition function. The main workflow and mechanisms of the proposed method, FDD-MOEA, are illustrated in Section \ref{sec:main}. Then, various empirical studies on test suites are given in Section \ref{sec:studies}. Finally, Section \ref{sec:conclusion} summarizes the contributions and future directions of this paper.

\section{Background and Motivations}
\label{sec:related}
In this part, we briefly illustrate the concept of multi-objective optimization, RBFN surrogate, federated learning and acquisition function. 

\subsection{Multi-objective optimization}
Denote the minimization of vector of objectives by 
\begin{equation}
    \begin{aligned}
    &\min_x \bm{f}(\bm{x}) = \left\{f_1(\bm{x}), f_2(\bm{x}), f_3(\bm{x}),...,f_M(\bm{x})\right\}^T, \\
    & \text{subject to} \quad \bm{x} \in R^d
    \end{aligned}
\end{equation}
where $\bm{x} \in R^d$ is the decision vector with dimension $d$, $M$ is the number of objectives and $M\ge 2$, and the profile formed by the projections of the aforementioned Pareto optimal solutions in the objective space is called Pareto front (PF). Note that this work focuses on the optimization of unconstrained MOPs.

\subsection{Radial basis function networks}
Radial basis function networks (RBFNs) \cite{du2014radidal} are widely adopted surrogate models due to their high computational efficiency and robustness on high dimensional problems. The first hidden layer connected to the input of an RBFN is composed by a series of centers $\{\bm{c}_1,...,\bm{c}_i,...,\bm{c}_q\}$, where $q$ is the number of centers, and $\bm{c}_i \in R^d$ is the $i^{th}$ center. This layer calculates the distance between a sample point $\bm{x} \in R^d$, $\phi \left(||\bm{x}-\bm{c}_i||\right)$, where $\phi(\cdot)$ represents a radial basis function, and in this work we use Gaussian functions with a spread $\delta_i$,
\begin{equation}
    \label{eq:gaussian}
    \phi \left(||\bm{x}-\bm{c}_i||\right)=e^{-\frac{||\bm{x}-\bm{c}_i||}{2\delta_i^2}}.
\end{equation}

Suppose $X\in R^{n_k\times d}$ and $Y\in R^{n_k\times M}$ represent a dataset with $n_k$ (indexing by $j$) sample pairs ($\bm{x}_j\in R^d, \bm{y}_j \in R^M$), then the corresponding output of the RBFN can be calculated by
\begin{equation}
	\label{eq:rbf_y}
	\resizebox{.99\hsize}{!}{$\begin{bmatrix}
	    \hat{y}_{1,1} &...& \hat{y}_{1,M} \\
	    &... & \\
	    \hat{y}_{n_k,1} &... &\hat{y}_{n_k,M} \\
	\end{bmatrix}=
    \begin{bmatrix}
	    \phi(||\bm{x}_1-\bm{c}_1||)& ...& \phi(||\bm{x}_1-\bm{c}_q||) \\
	    & ... & \\
        \phi(||\bm{x}_{n_k}-\bm{c}_1||)& ...& \phi(||\bm{x}_{n_k}-\bm{c}_q||) \\
	\end{bmatrix}
	\begin{bmatrix}
	    w_{1,1} &... & w_{1,M}  \\
	    &... &\\
	    w_{q,1} &... & w_{q,M} \\
	\end{bmatrix} +
	\begin{bmatrix}
	    b_{1} &...b_{M}  \\
	\end{bmatrix}$},
\end{equation}
which can further be simplified to be
\begin{equation}
    \hat{\bm{y}}_j = \Phi_j \bm{w} + \bm{b}, \; \text{for}\; j=1,2,...,n_k
\end{equation}
where $\Phi_j=[\phi(||\bm{x}_j-\bm{c}_1||),...,\phi(||\bm{x}_j-\bm{c}_q||)]$ denotes the distance between sample $x_j$ and all centers, and $\bm{w}=[w_{1},..,w_{M}]\in R^{q\times M}$, in which $w_{M}=[w_{1,M},w_{2,M},...,w_{q,M}]^T$ are the weights of the connecting $M^{th}$ attribute of $\bm{y}_j$ and all centers, and $\bm{b}=[b_1,...,b_M]\in R^{1\times M}$ is the bias, $\bm{\delta}=\{\delta_1,\delta_2,...,\delta_q\}\in R^q$ stands for the spread vector of the RBFN. Hence, we can use $[\bm{w},\bm{\delta},\bm{c},\bm{b}]$ to represent the RBFN.

In terms of the training process, the centers of an RBFN is usually determined by random selection or K-means clustering. The least square method and backpropagation are two most popular methods \cite{du2014radidal} to train the weights $\bm{w}$ and biases $\bm{b}$. The least square method optimizes the weights and the biases through matrix inversion (linear algebra), whereas backpropagation uses strategies such as the stochastic gradient descent (SGD) method \cite{robbins1951stochastic} to iteratively optimize the parameters.

Compared with GPs, RBFNs are efficient for training and it is less sensitive to the increase of training data size. However, an RBFN cannot directly provide the uncertainty of its predictions like GPs, making it unsuitable for applying an infill criterion for surrogate model management.

\subsection{Federated learning}
\label{sec:FL}
Federated learning \cite{mcmahan2017communication} originates from the attempt to address the privacy concerns in distributed learning \cite{li2014scaling}, which has been widely studied and applied in real-world applications due to its capability of privacy protection and parallel computing \cite{xu2020ternary,zhu2021federated}. Generally, the global optimization objective of a federated learning system can be written by
\begin{equation}
    \label{eq:global_loss}
    \min_{W_s} F(W_s) = \sum_{k=1}^{\lambda N}p_k F_k(W_k),
\end{equation}
where $N$ is the total number of clients, $\lambda$ is a coefficient that reflects how many clients will participate in model update in each communication round, and $p_k$ is the weight that controls the importance of the participating client $k$, which is usually determined by the size of local dataset $D_k$,
\begin{equation}
    \label{eq:pk}
    p_k = \frac{|D_k|}{\sum_{k=1}^{\lambda N}|D_k|},
\end{equation}
and $F_k$ denotes the local objective, which is
\begin{equation}
    \label{eq:local_loss}
    F_k(W_k) = \sum_{i=1}^{n_k} L(W_k;\bm{x}_i,\bm{y}_i).
\end{equation}
where $L$ is the loss function.

As Mcmahan \cite{mcmahan2017communication} suggested, the global model $W_s$ can be conducted by averaging a set of local models $W_k$ $\left(k=\{1,2,...,\lambda N\}\right)$ rather than collecting distributed local data, which can be denoted by
\begin{equation}
    \label{eq:fedavg}
    W_s = \sum_{k=1}^{\lambda N} p_k W_k.
\end{equation}

To summarize, federated learning is a distributed learning framework, in which a local model is trained on each client. The trained local models are then uploaded onto the server and aggregated into a global model. The global model is then downloaded to all participating clients to be further trained. Ideally, the global model will converge to the one that is trained on all local data that are centrally stored.  


It should be emphasized that the main motivation of the present work is to achieve data-driven optimization of complex systems, in which the data is collected and distributed on multiple clients. Due to privacy concerns and other considerations, the data is not allowed to transmitted to a central server. To the best of our knowledge, all existing work on data-driven evolutionary optimization assumes that the sampled data are available for training a global surrogate model using centralized learning, with one exception reported very recently in \cite{xu2021federated}, which proposed, for the first time, an algorithm for federated data-driven single-objective optimization. Actually, this work is a natural extension of \cite{xu2021federated}, aiming to propose a federated data-driven evolutionary multi-objective optimization algorithm, FDD-MOEA for short.  


\subsection{Acquisition function}
\label{sec:infill}
Acquisition function (AF), also called infill criterion or model management, is designed to decide where to sample new data (new candidate solutions) in the decision space, i.e., to determine which candidate solutions $\bm{x}$ should be selected for calculating their objective values $\bm{f}(\bm{x})$, usually by means of time-consuming numerical simulations or physical experiments. Acquisition functions were originally developed in Bayesian optimization when the Gaussian process is used as the surrogate, although it can be easily extended to any other types of surrogates with the capability of providing uncertainty information, such as ensemble surrogates \cite{guo2018heterogeneous}, or dropout neural networks \cite{guo2021evolutionary}. 

An infill criterion must be able to balance the exploitation and exploration \cite{jones1998efficient}, and several successful infill criteria have been widely adopted by surrogate-assisted MOEAs, such as Expectation Improvement (EI) \cite{jones1998efficient} and Lower Confidence Bound (LCB)  \cite{torczon1998using}. In this work, we choose LCB to minimize Eq. (\ref{eq:LCB}), which is
\begin{equation}
    \label{eq:LCB}
    LCB(\bm{x}) = \hat{\bm{f}}(\bm{x})-\alpha \hat{s}(\bm{x}),
\end{equation}
where $\alpha$ is a hyperparameter which is set to 2 as suggested in \cite{emmerich2006single} in the experimental studies.



\section{Federated Data-driven Multi-objective Optimization}
\label{sec:main}
In this section, we will briefly introduce the federated construction of the global model, the workflow of the proposed FDD-MOEA, and the used federated infill criterion.

\subsection{Federated surrogate construction}
\label{sec:surrogate_train}

\textbf{Clients}: Suppose $W_k=[\bm{w}_k,\bm{c}_k,\bm{\delta}_k,\bm{b}_k]$ represents the surrogate on the $k^{th}$ client, and let $\{\hat{\bm{y}}_1,\hat{\bm{y}}_2,...,\hat{\bm{y}}_{B}\}$ represent the predicted outputs with a batch size of $B$ ($1\le B \le n_k$), we have
\begin{equation}
    \label{eq:y_hat}
    \hat{\bm{y}}_i=W_k(\bm{x}_i)= \Phi_i \bm{w} + \bm{b}, \; \text{for}\; i=1,2,...,B  
\end{equation}
then the loss function ($L$ in Eq. (\ref{eq:local_loss})) of each local RBFN used in this work can be written as
\begin{equation}
    \label{eq:mse}
    L=\frac{1}{2B}\sum_{i=1}^{B}(\hat{\bm{y}}_i-\bm{y})^2,
\end{equation}
and the centers and spreads of the RBFN are determined by K-means maximum distance between centers, then the gradients of weights $\bm{w}_k$ and $\bm{b}_k$ are calculated by
\begin{equation}
    \label{eq:gradient_w}
    \frac{\partial  L}{\partial  \bm{w}_k} = \frac{1}{B}\sum_{i=1}^B \frac{\partial \hat{\bm{y}}_i}{\partial \bm{w}_k}(\hat{\bm{y}}_i-\bm{y}),
\end{equation}

\begin{equation}
    \label{eq:gradient_b}
    \frac{\partial  L}{\partial  \bm{b}_k} = \frac{1}{B}\sum_{i=1}^B \frac{\partial \hat{\bm{y}}_i}{\partial \bm{b}_k}(\hat{\bm{y}}_i-\bm{y}).
\end{equation}

By replacing $\hat{\bm{y}}_i$ with Eq. (\ref{eq:y_hat}), we have
\begin{equation}
    \label{eq:gradient_w2}
    \frac{\partial  L}{\partial  \bm{w}_k} = \frac{1}{B}\sum_{i=1}^B \Phi_i(\hat{\bm{y}}_i-\bm{y}),
\end{equation}

\begin{equation}
    \label{eq:gradient_b2}
    \frac{\partial  L}{\partial  \bm{b}_k} = \frac{1}{B}\sum_{i=1}^B (\hat{\bm{y}}_i-\bm{y}).
\end{equation}

\begin{figure}[htpb]
    \centering
    \includegraphics[width=0.75\textwidth]{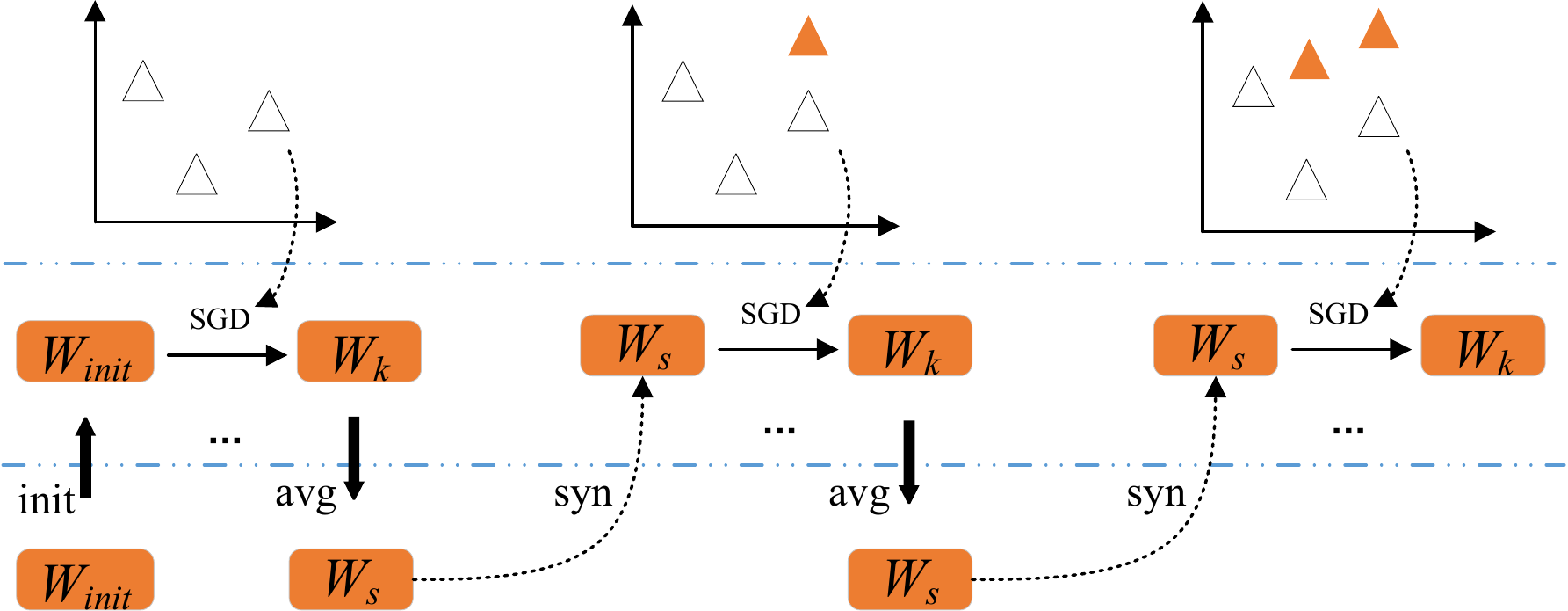}
    \caption{An illustration of local training with server model.}
    \label{fig:train_examp}   
\end{figure}

In this way, each client iteratively trains the weights and biases of the local $W_k$ with a learning rate $\eta$ and a local epoch $E$. In addition, the batch size $B$ is set to 1 in training the local models in this work, mainly because the number of training data on each client is very limited in data-driven optimization. 

\textbf{Server}: After the local model on the $k$-th client ($W_k$) is obtained and uploaded to the server, the sorted averaging algorithm, as described in Algorithm \ref{algo:sorted_avg} and first reported in \cite{xu2021federated}, will be adopted to aggregate all received local models to create the global model $W_s$. Then, the server will use $W_s$ to assist an MOEA. Finally, $W_s$ and the found potential solution $\bm{x}$ to the chosen clients, the $k^{th}$ client will evaluate the $\bm{x}$ by the expensive objectives and update the local dataset to further update local model $W_k$, which can be illustrated by Fig. \ref{fig:train_examp}. The overall framework will be described in Section \ref{sec:framework}.

\begin{algorithm}[H]

	\caption{Sorted Averaging \cite{xu2021federated}}
	\label{algo:sorted_avg}
	\SetKwInOut{Input}{Input}
	\SetKwInOut{Output}{Output}
	\Input{Local surrogates $W_k=[\bm{w}_k,\bm{c}_k,\bm{\delta}_k,\bm{b}_k]$, global surrogate $W_s$.}
	\vspace{3pt}

	\ForEach{k {\rm= 1, 2, 3 ... $\lambda N$}}
	{

		\vspace{3pt}
		index $\leftarrow$ Sort according to the distance of each center in $\bm{c}_k$ from the origin 
		
		\vspace{3pt}
		$\bm{c}_k$, $\bm{w}_k$, $\bm{\delta}_k$ $\leftarrow$ $\bm{c}_k$[index], $\bm{w}_k$[index], $\bm{\delta}_k$[index]
		
		\vspace{3pt}
	     $W_k = [\bm{c}_k; \; \bm{w}_k; \; \bm{\delta}_k; \; \bm{b}_k]$
	}
	\vspace{3pt}
	
	$ W_s \leftarrow \sum_{k=1}^{\lambda N}p_k W_k$
	\vspace{6pt}
	
	\Output{Server model $W_s$}
\end{algorithm}

\subsection{Overall framework}
\label{sec:framework}
\begin{figure*}[htbp]
    \centering
    \includegraphics[width=0.95\textwidth]{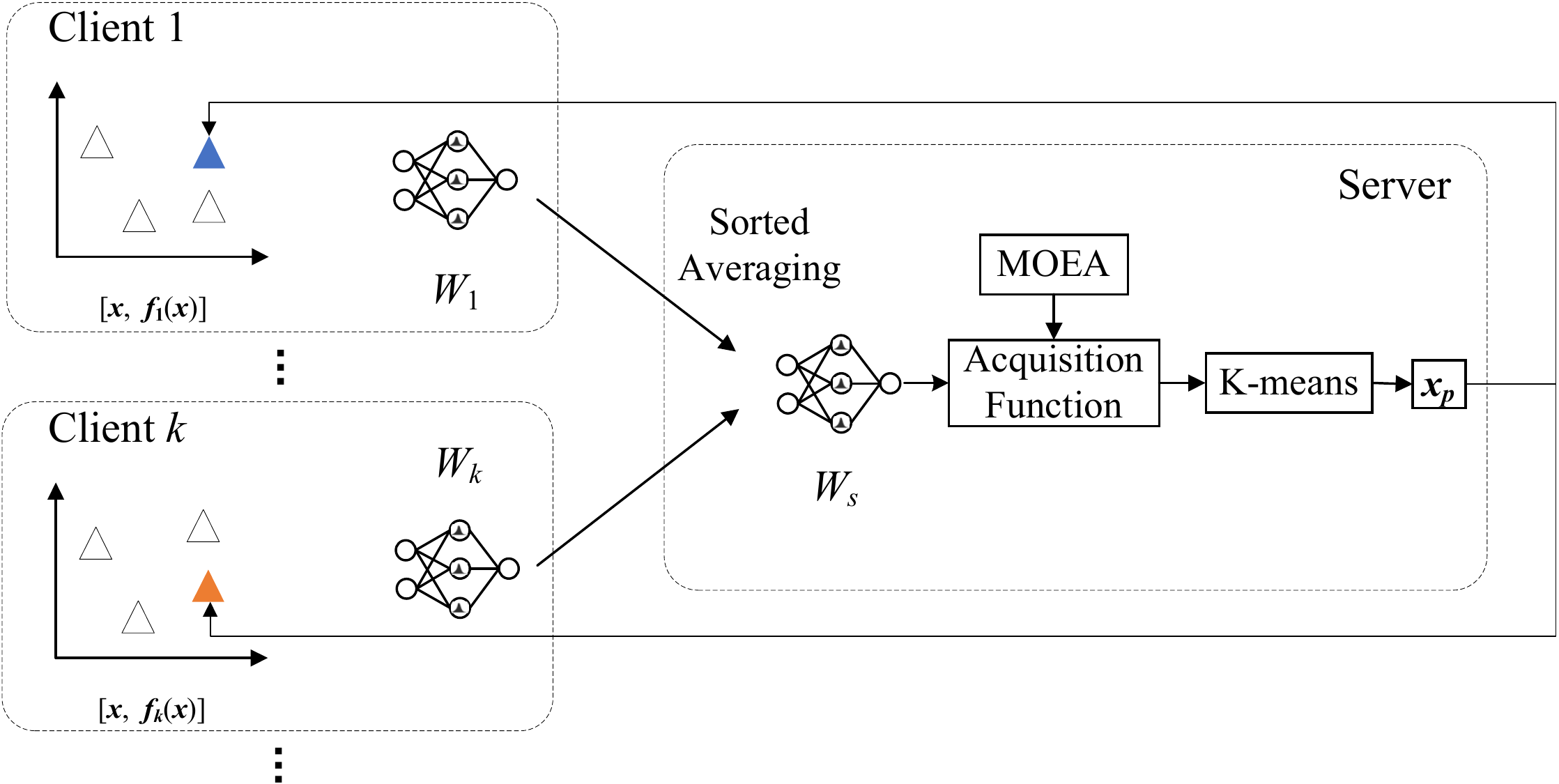}
    \caption{The overall framework of the proposed FDD-MOEA.}
    \label{fig:framework}   
\end{figure*}

As shown in Fig. \ref{fig:framework}, the proposed FDD-MOEA differs from those of ensemble-based methods in that it utilizes sorted model averaging to obtain a global model $W_s$ for fitness approximation instead of averaging predictions of ensemble. In addition, the data partitions for each local models are determined by the operating conditions of the clients and cannot be actively manipulated as done conventional ensemble construction when the data is centrally stored. It also be stress that in principal, any MOEAs can be applied to the proposed framework. In our experimental studies, we choose NSGA-II and RVEA to examine the performance of the proposed framework for MOPs and MaOPs, respectively.

Algorithm \ref{algo:FDD-MOEA} summarizes the overall workflow of the proposed FDD-MOEA. Before the optimization starts, Latin hypercube sampling (LHS) is adopted to generate $11d-1$ points in the decision space, where $d$ is the dimension of the decision space. And these initial points are evaluated by the expensive real objective functions (data collection), the collected data pairs are used as the initial training dataset of for the local models. Then, the $k^{th}$ client will train the local RBFN $W_k$ with its own dataset and upload the trained $W_k$ to the server. The server aggregates all uploaded local models by means of sorted averaging to obtain the global model $W_s$. After that, the acquisition function (AF) will be formed using $W_s$ and all local models (refer to Section \ref{sec:AF}), in which the averaged $W_s$ will replace the expensive objective functions to predict the objective values, and the local models will collaboratively estimate the uncertainty of the predictions.

\begin{algorithm}
	\label{algo:FDD-MOEA}
	\caption{Federated data-driven MOEA.}
	\SetKwInOut{Input}{Input}
	\SetKwInOut{Output}{Output}
	\SetKwProg{Server}{Server does:}{}{end}
	\Input{Dimension of the decision space $d$; Expensive real objective functions $\bm{f}$; Maximum iterations $R$; Maximum number of training samples $l$; Number of clients $N$; Number of selected samples per round $m$; Failure probability of the communication $p_f$.}
    \vspace{3pt}
	\Output{Solutions $X$ and their objective values $Y$}
	\vspace{3pt}

    $X\leftarrow$ Latin hypercube sampling $11d-1$ points

    \vspace{3pt}

    \For{$i$ {\rm= 1 to $11d-1$}}{

        \vspace{3pt}
        $Y[i]\leftarrow \bm{f}(X[i])$
    }
	\vspace{3pt}
	\For{r {\rm= 1, 2, 3 ... $R$}}
	{
        \vspace{3pt}
        \For{{\bf Client} {\rm= 1, 2, 3 ... $\lambda N$ in parallel}}
        {
            \vspace{3pt}
            // communication succeeded ($1-p_f$)
            
            \vspace{3pt}
            evaluate $Y_{new}$ by $\bm{f}$
        
            \vspace{3pt}
        	$X,\; Y \leftarrow X\cup X_{new}, Y \cup Y_{new}$
        	
            \vspace{3pt}
            \If{$|X|>l$}{

                \vspace{3pt}
                $X_k, Y_k \leftarrow$ select($X,\;Y,\;l$)     \quad // fast non-dominated sorting
            }

            \vspace{3pt}
            $W_k \leftarrow $ local training ($X_k, Y_k$)
        }
    
    \Server{}
        {
        \vspace{3pt}
    	$W_s \leftarrow$ Algorithm \ref{algo:sorted_avg}
    	
    	\vspace{3pt}
    	AF $\leftarrow$ Modified LCB (Section \ref{sec:AF})
    
        \vspace{3pt}
    	solutions $\leftarrow$ MOEA(AF)
    
        \vspace{3pt}
        $X_{new} \leftarrow$ select $m$ new solutions
        }
    }
\end{algorithm}

Then, an MOEA is chosen to optimize the acquisition functions, and $m$ new solutions (specified by the user) will be selected from the results of MOEA through clustering. Note, however, that duplicate solutions will be removed before clustering.
\begin{itemize}
    \item[1)] Removal of redundant solutions. It is very common when the optimization completes, the final population of the MOEA contains redundant solutions or extremely similar solutions. In this work, we calculate the Euclidean distances between solutions of the current population in the decision space and only one will be kept if the distance between two solutions is less than a threshold ($10^{-6}$ in this work). Note that the number of solutions after removing the redundant solutions may be smaller than $m$, in particular for high-dimension problems. In this case,  we will save the existing solutions and restart the MOEA until the stopping criterion is satisfied.
    \item[2)] Clustering. After removing the redundant solutions, the remaining ones will be clustered using the K-means algorithm with a center size of $m$. 
\end{itemize}

In federated optimization, there are many challenges that may lead to difficulties in collecting well distributed data, making it hard to achieve a high-quality global model. In this work, we consider two main challenges, namely communication failures and client dropout. We define a probability of communication failure $p_f$ to denote whether the communication between a certain client and the server is successful in one particular iteration of surrogate management. If the communication is successful, then all $m$ candidate solutions selected by the server will be passed to the participating clients, evaluated, and added to the local training dataset for the next round training. Otherwise, the $m$ new solutions will not be evaluated by that particular client. Consequently, the larger the communication failure probability, the stronger the data will be non-independent and identically distributed (non-iid). 

Client dropout is also caused by communication problems. Like in federated learning, we use a participation ratio $\lambda$ to define the percentage of clients that participate in each round of model update and surrogate management. The smaller participation ratio, the smaller the number of local models, which may result in a poorer global model.   

To be consistent with the settings in \cite{guo2018heterogeneous}, we also limit the maximum number of training data to be $l=11d-1+25$. A fast non-dominated sorting is applied to select $l$ data for model training.

\subsection{Modified LCB}
\label{sec:AF}
As we stated in Section \ref{sec:infill}, the acquisition function prefers solutions with high uncertainty and better performance. To achieve this, we utilize the global surrogate model together with the local models to provide predictions of candidate solutions and estimate the uncertainty of the predictions. 

First, once the server model $W_s$ is aggregated, the evaluated objective values $\hat{f}(\bm{x})$ in Eq. (\ref{eq:LCB}) is calculated by
\begin{equation}
	\label{eq:f_mean}
	\hat f(\bm{x}) = W_s(\bm{x}).
\end{equation}
Similar to the ensemble-based strategy, the variance $\hat s^2(\bm{x})$ of the predictions can be estimated by
\begin{equation}
    \label{eq:s2}
	\hat s^2(\bm{x}) = \frac{1}{\lambda N-1}
\left[\sum_k^{\lambda N}\left(W_k(\bm{x}) - \hat f(\bm{x})\right)^2 \right].
\end{equation}
Consequently, we can easily calculate the LCB values for optimization using the MOEA. 

\section{Numerical Studies}
\label{sec:studies}
In the following, the benchmark problems and parameter settings used in the empirical studies are introduced and then the comparative results are presented.

\subsection{Benchmark problems}
As listed in Table \ref{Tab:benchmark}, we adopt the DTLZ test suite for comparative studies, and 49 test instances in total with different number of decision variables and objectives are used. According to \cite{huband2006review}, an MOEA may get trapped in local Pareto front of DTLZ1 and DTLZ3, which will lead to poor performance. And the PF of DTLZ2 are located in an octant of the unit sphere, which is often used to test the performance of an MOEA with a large number of objectives. It is hard for an MOEA to find evenly distributed solutions in the  objective space for DTLZ4 due to the biased projection from decision space to the objective space \cite{guo2018heterogeneous}, and the mapping parameter $\alpha$ of DTLZ4 is set to 100 in this work. DTLZ5 and DTLZ6 have degenerate Pareto optimal fronts, and each front is supposed to be an arc in the objective space. The Pareto front of DTLZ7 is disconnected.

\begin{table}
	\caption{Test instances and their properties.}
	
	\label{Tab:benchmark}
	\centering
	\renewcommand\arraystretch{1.8}
	\setlength{\tabcolsep}{2.5mm}
    \begin{tabular}{c|c|c|c|c}
    \hline
    Problem  & $M\;(d=20)$    & $d\;(M=3)$     & Domain   & Pareto front         \\ \hline
    DTLZ1 & 3, 5, 10, 20  & 10, 20, 40, 80  & [0, 1]   & Linear           \\ \hline
    DTLZ2 & 3, 5, 10, 20  & 10, 20, 40, 80  & [0, 1]   & Concave          \\ \hline
    DTLZ3 & 3, 5, 10, 20  & 10, 20, 40, 80  & [0, 1]   & Concave, biased  \\ \hline
    DTLZ4 & 3, 5, 10, 20  & 10, 20, 40, 80  & [0, 1]   & Concave, biased          \\ \hline
    DTLZ5 & 3, 5, 10, 20  & 10, 20, 40, 80  & [0, 1]   & Degenerate       \\ \hline
    DTLZ6 & 3, 5, 10, 20  & 10, 20, 40, 80  & [0, 1]   & Degenerate       \\ \hline
    DTLZ7 & 3, 5, 10, 20  & 10, 20, 40, 80  & [0, 1]   & Disconnected     \\ \hline
    \end{tabular}
\end{table}

\subsection{Parameter settings}

Recall that the number of objectives and the dimension of the decision variable are denoted by $M$ and $d$, respectively. The basic parameter settings are as follows.
\begin{enumerate}
    \item The number of independent runs is set to 20 for each test instance, and each time we sample $11d\;-\;1$ points in the decision space using LHS.
    
    \item NSGA-II is chosen as the MOEA for MOPs, the population size and the number of generations are both set to 50. The crossover distribution index and its probability are set to 20, 1.0. And the corresponding mutation distribution index and probability are 20 and $1/d$.
    
    \item RVEA is adopted as the MOEA for MaOPs, and the number of reference vectors are set to 105, 126, 275, 420 for 3, 5, 10, 20 objectives, respectively.
    
    \item As suggested in \cite{guo2018heterogeneous}, the number of RBFN centers is limited to $\sqrt{M+d}\;+\;3$, the optimization process starts at $11d-1$ fitness evaluations (FEs) and ends at $11d-1+120$ FEs, and five new FEs are performed (five new training samples are collected) in each round, which means that the number of iteration $R$ is 24 in a single run. The maximum size of training data $l$ is set to $11d-1+25$. 
    
    \item The hyperparameters of federated learning are set as follows: the local epoch number $E=20$, the learning rate $\eta=0.06$, the number of clients $N=10$, and the participation ratio $\lambda=0.9$. As previously discussed, for each individual client, the solutions to be sampled at each iteration as identified by the surrogate management strategy by the server may not be implemented on each client with a communication failure probability $p_f=3\%$. To make a fair comparison with ensemble based algorithms, the initial solutions are the same on all clients.
    
    \item Inverted generational distance (IGD) \cite{bosman2003balance} is adopted as the performance indicator, which calculates the Euclidean distance between the non-dominated solutions obtained by an algorithm and a set of reference points sampled from the theoretical Pareto front. The result of 20 independent runs are analyzed by the Wilcoxon rank sum at a significance level of 0.05, where symbol '$+$' means that FDD-MOEA outperforms the compared method, '$=$' indicates the two algorithms perform equally well, and '$-$' means that the compared method outperforms FDD-MOEA.
    
\end{enumerate}

\subsection{Results and discussions}
We implement the proposed FDD-MOEA with Python 3.7 \cite{van1995python}, and the RBFN is constructed from scratch using the stochastic gradient descent method. The code of NSGA-II is taken from Pymoo \cite{blank2020pymoo}, and the K-RVEA are implemented by PlatEMO v2.8.0 \cite{tian2017platemo}, and the performance indicators are also calculated on PlatEMO. All experiments are conducted on a computer with Intel(R) Core(TM) i7-8700 CPU@3.20GHz 3.19 GHz, Windows 10, version 1909.

\begin{table}
	\caption{Mean (std) of the IGD values obtained by FDD-MOEA, HeE-MOEA, GP-MOEA and K-RVEA for different decision dimensions when $M=3$.}
	
	\label{Tab:d}
	\centering
	\renewcommand\arraystretch{1.8}
    \begin{tabular}{c|c|c|c|c|c}
    \hline
    Problems   & d     & FDD-MOEA        & HeE-MOEA         & GP-MOEA   & K-RVEA                   \\ \hline
    
    \multirow{4}{*}{DTLZ1}
    & 10  & 101.71 (2.0e+01)  & 111.95 (2.2e+01) =  & 116.18 (1.4e+01) = &\cellcolor{gray!50}92.962 (2.2e+01) =  \\ 
    & 20  &\cellcolor{gray!50}327.79 (3.8e+01)  & 356.33 (4.3e+01) =  & 363.78 (3.5e+01) + & 330.06 (4.7e+01) = \\ 
    & 40  &\cellcolor{gray!50}864.74 (5.5e+01)  & 900.74 (5.4e+01) +  & 896.17 (4.0e+01) + & 887.47 (4.4e+01) = \\ 
    & 80  &\cellcolor{gray!50}1949.2 (8.5e+01)  & 2078.1 (6.5e+01) +  & 1991.7 (6.9e+01) = & 2028.0 (7.9e+01) + \\ \hline
    
    \multirow{4}{*}{DTLZ2}
    & 10  &\cellcolor{gray!50}0.1738 (1.9e-02)  & 0.1958 (2.0e-02) +  & 0.2028 (7.1e-02) + & 0.1829 (3.6e-02) =  \\
    & 20  &\cellcolor{gray!50}0.2969 (3.1e-02)  & 0.3261 (2.1e-02) +  & 0.5353 (6.1e-02) + & 0.6032 (7.5e-02) + \\ 
    & 40  &\cellcolor{gray!50}0.4214 (2.7e-02)  & 0.7913 (6.6e-02) +  & 0.9432 (1.0e-01) + & 1.7960 (1.3e-01) + \\ 
    & 80  &\cellcolor{gray!50}0.6889 (3.2e-01)  & 2.1337 (3.1e-01) +  & 1.9965 (1.7e-01) + & 4.6585 (2.1e-01) + \\ \hline
    
    \multirow{4}{*}{DTLZ3} 
    & 10  & 285.16 (5.7e+01)  & 315.84 (6.3e+01) =  & 330.81 (5.7e+01) + &\cellcolor{gray!50}236.74 (4.3e+01) - \\ 
    & 20  & 1056.8 (7.6e+01)  & 1073.5 (1.2e+02) =  & 1094.8 (9.1e+01) + &\cellcolor{gray!50}941.95 (1.3e+02) - \\ 
    & 40  &\cellcolor{gray!50}2734.7 (1.4e+02)  & 2807.9 (1.0e+02) +  & 2810.3 (1.6e+02) + & 2776.3 (1.7e+02) = \\ 
    & 80  &\cellcolor{gray!50}6323.1 (3.4e+02)  & 6363.2 (2.3e+02) =  & 6366.2 (3.1e+02) = & 6382.9 (1.9e+02) = \\ \hline
    
    \multirow{4}{*}{DTLZ4} 
    & 10  & 0.7199 (9.5e-02)  & 0.8060 (1.1e-01) =  & 0.6583 (7.6e-02) - &\cellcolor{gray!50}0.4139 (1.3e-01) -   \\ 
    & 20  & 0.9331 (2.2e-01)  & 0.9614 (2.5e-02) +  & 1.0231 (1.1e-01) + &\cellcolor{gray!50}0.8715 (1.4e-01) =  \\ 
    & 40  &\cellcolor{gray!50}0.9823 (1.1e-02)  & 1.2422 (4.5e-02) +  & 1.9227 (1.4e-01) + & 2.4490 (1.0e-01) + \\ 
    & 80  &\cellcolor{gray!50}1.3043 (4.8e-02)  & 2.5379 (1.4e-01) +  & 2.7526 (6.2e-02) + & 5.0597 (1.5e-01)  + \\ \hline
    
    \multirow{4}{*}{DTLZ5} 
    & 10  &\cellcolor{gray!50}0.0604 (1.3e-02)  & 0.1125 (2.0e-02) +  & 0.1466 (2.6e-02) + & 0.1067 (3.6e-02) + \\
    & 20  &\cellcolor{gray!50}0.1528 (2.7e-02)  & 0.2060 (3.5e-02) +  & 0.4607 (1.0e-01) + & 0.4018 (8.0e-02) +  \\
    & 40  &\cellcolor{gray!50}0.2643 (3.1e-02)  & 0.6681 (4.8e-02) +  & 0.9121 (1.4e-01) + & 1.6052 (1.4e-01) +  \\ 
    & 80  &\cellcolor{gray!50}0.5754 (3.2e-02)  & 1.9576 (1.7e-01) +  & 1.7653 (2.0e-01) + & 4.5680 (2.5e-01) +  \\ \hline
    
    \multirow{4}{*}{DTLZ6} 
    & 10  & 6.7366 (7.2e-02)  & 5.8677 (4.3e-01) -  & 5.7072 (3.2e-01) - &\cellcolor{gray!50}3.2101 (4.9e-01) - \\
    & 20  & 15.394 (1.4e-01)  & 15.037 (4.0e-01) -  & 13.158 (3.7e-01) - &\cellcolor{gray!50}11.016 (9.6e-01) - \\
    & 40  & 32.312 (3.4e-01)  & 32.561 (6.8e-01) =  & 32.194 (6.8e-01) = &\cellcolor{gray!50}30.688 (7.6e-01) -  \\
    & 80  &\cellcolor{gray!50}66.721 (2.3e-01)  & 67.808 (7.4e-01) =  & 68.032 (6.2e-01) + & 67.980 (7.1e-01) =  \\ \hline
    
    \multirow{4}{*}{DTLZ7} 
    & 10  & 5.9812 (9.8e-01)  & 1.8227 (6.3e-01) -  & 0.9151 (3.2e-01) - &\cellcolor{gray!50}0.1570 (5.7e-02) - \\ 
    & 20  & 7.9733 (7.1e-01)  & 4.4650 (7.6e-01) -  & 4.3612 (7.1e-01) - &\cellcolor{gray!50}0.3631 (1.7e-01) - \\ 
    & 40  & 9.3081 (9.4e-01)  & 6.5439 (7.2e-01) -  &\cellcolor{gray!50}5.8972 (1.0e+00) - & 6.4692 (2.0e+00) -  \\ 
    & 80  & 9.7757 (3.3e-01)  &\cellcolor{gray!50}8.7749 (3.5e-01) -  & 9.8455 (2.3e-01) = & 9.5087  (4.4e-01) = \\ \hline
    
    \multicolumn{2}{c|}{+/-/=}  & -        &  14/6/8  & 17/5/6   & 10/9/9 \\ \hline
    
    \end{tabular}
\end{table}

\subsubsection{Performance comparison on three-objective test instances}
\label{sec:result_d}
The first set of experiments compares FDD-MOEA with three centralized data-driven MOEAs, namely HeE-MOEA \cite{guo2018heterogeneous}, GP-MOEA \cite{guo2018heterogeneous,guo2021evolutionary} and K-RVEA \cite{chugh2016surrogate} on three-objective DTLZ functions with a dimension of the decision spacce $d=10,20,40,80$. Specially, HeE-MOEA utilizes a heterogeneous ensemble to assist the optimization, while GP-MOEA is a variant of HeE-MOEA that adopts a GP as the surrogate for approximating each objective function. Note that HeE-MOEA, GP-MOEA and FDD-MOEA use NSGA-II as the baseline optimizer, while K-RVEA is based on RVEA. The average IGD results on all these test instances are summarized in Table \ref{Tab:d}. 

From the results in Table \ref{Tab:d}, we can observe that FDD-MOEA obtains the best IGD value on 16 out of 28 test instances, compared with HeE-MOEA, GP-MOEA and K-RVEA. These results confirm that the proposed FDD-MOEA works effectively for solving MOPs in a federated system. We also note that K-RVEA performs better than HeE-MOEA and GP-MOEA. It should be stressed that that the purpose of proposing FDD-MOEA is not for developing a new MOEA that performs better than the state-of-the-art data-driven optimization algorithms. Rather, it aims to propose an algorithm that can still work effectively when the data are distributed on different clients and are not allowed to centrally stored. 

\begin{figure*}
    \centering
    \subfigure[\centering DTLZ1, $M=3$, $d=10$]{{\includegraphics[width=0.48\textwidth]{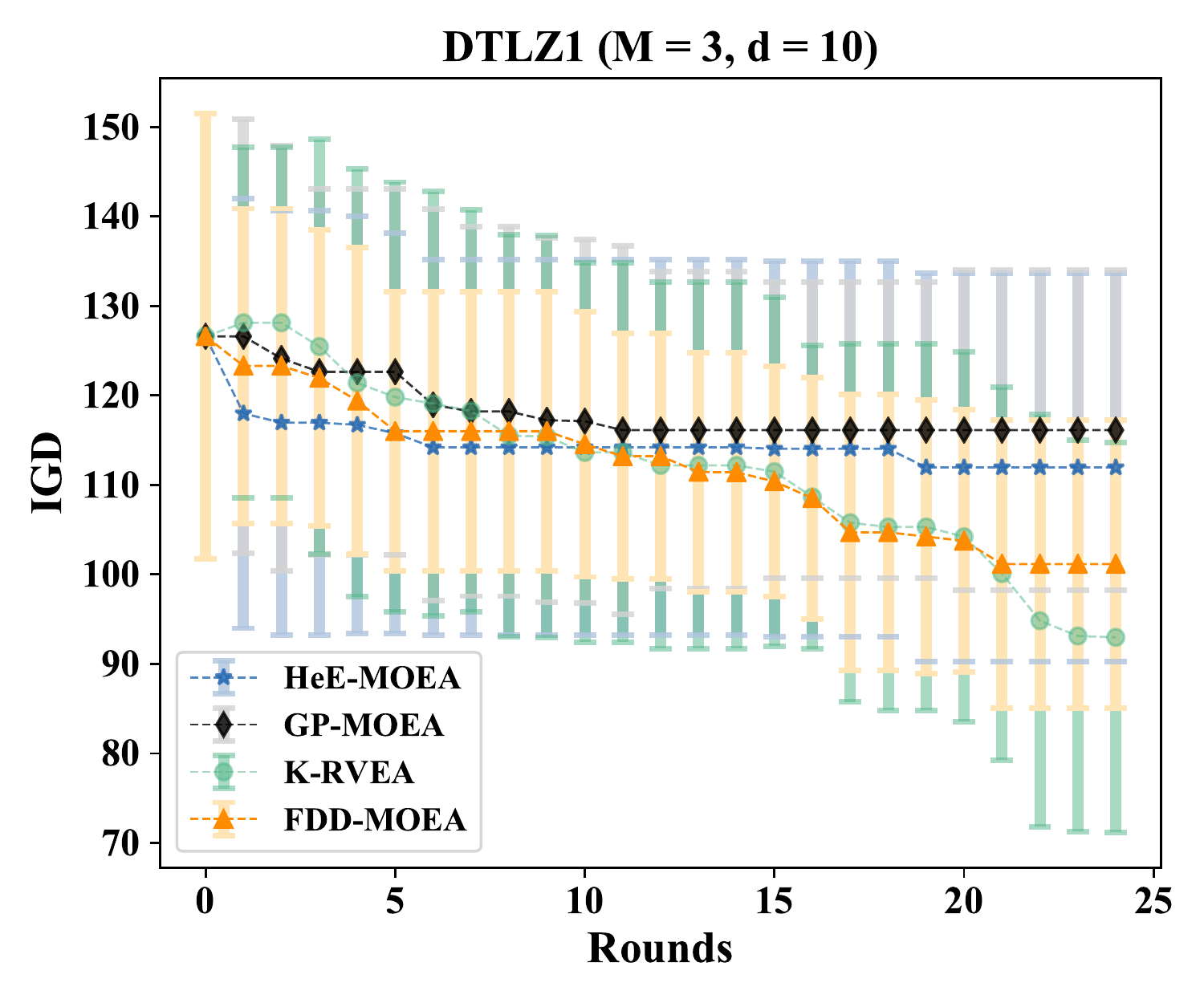} }}
    \subfigure[\centering DTLZ1, $M=3$, $d=80$]{{\includegraphics[width=0.48\textwidth]{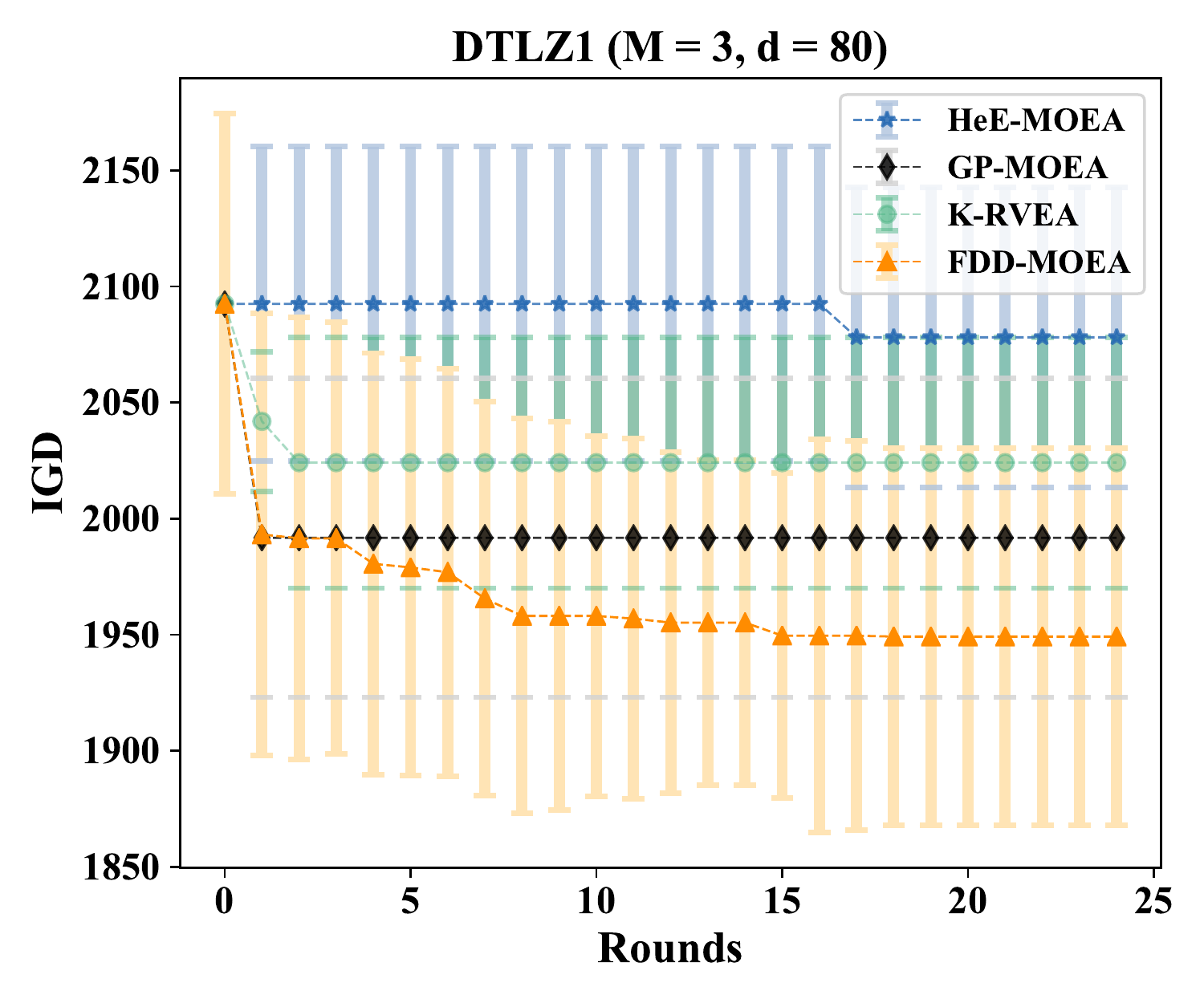} }}
    \caption{The IGD convergence profiles on DTLZ1.}
    \label{fig:DTLZ1}
\end{figure*}

\begin{figure*}
    \centering
    \subfigure[\centering DTLZ2, $M=3$, $d=10$]{{\includegraphics[width=0.48\textwidth]{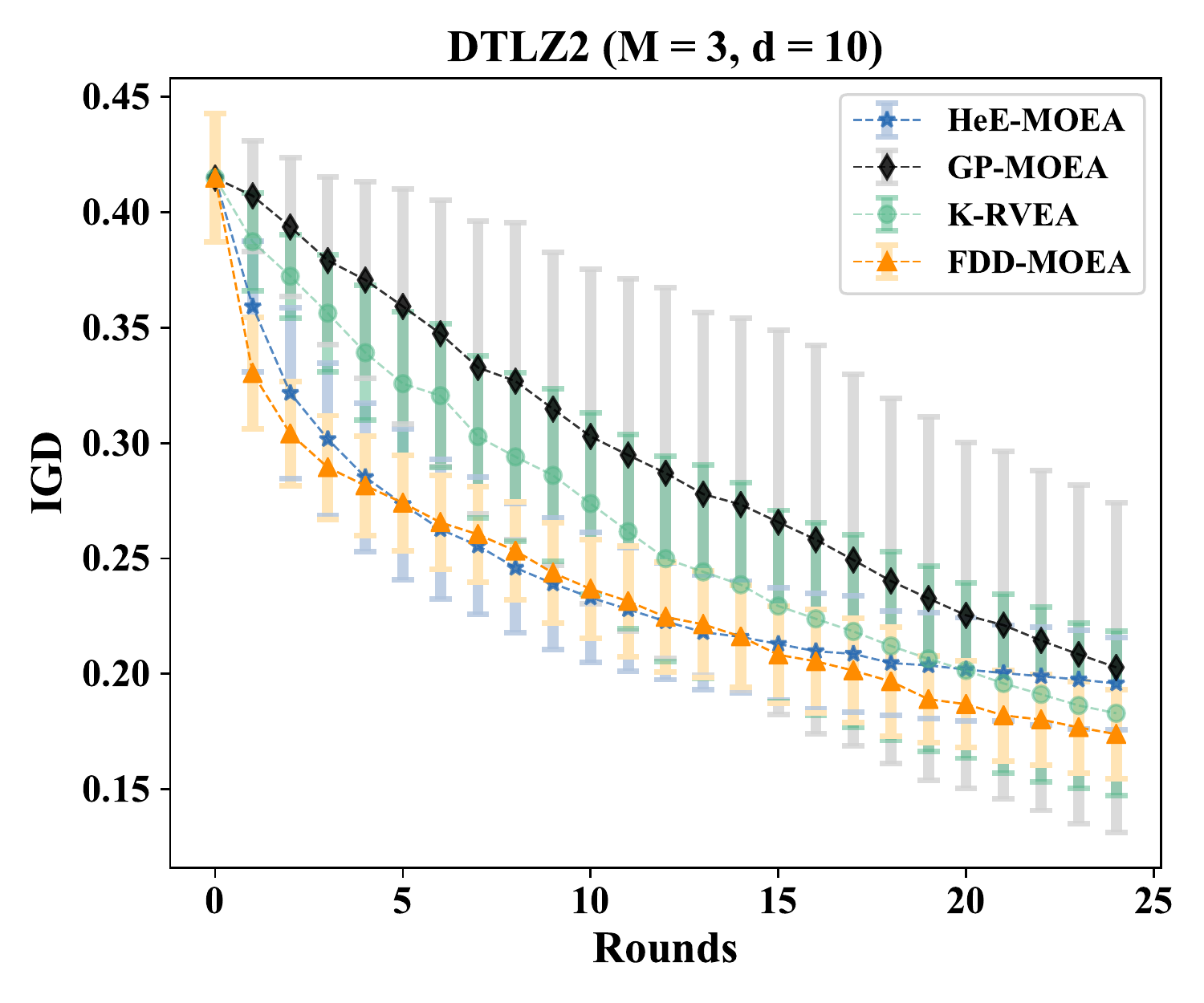} }}
    \subfigure[\centering DTLZ2, $M=3$, $d=80$]{{\includegraphics[width=0.48\textwidth]{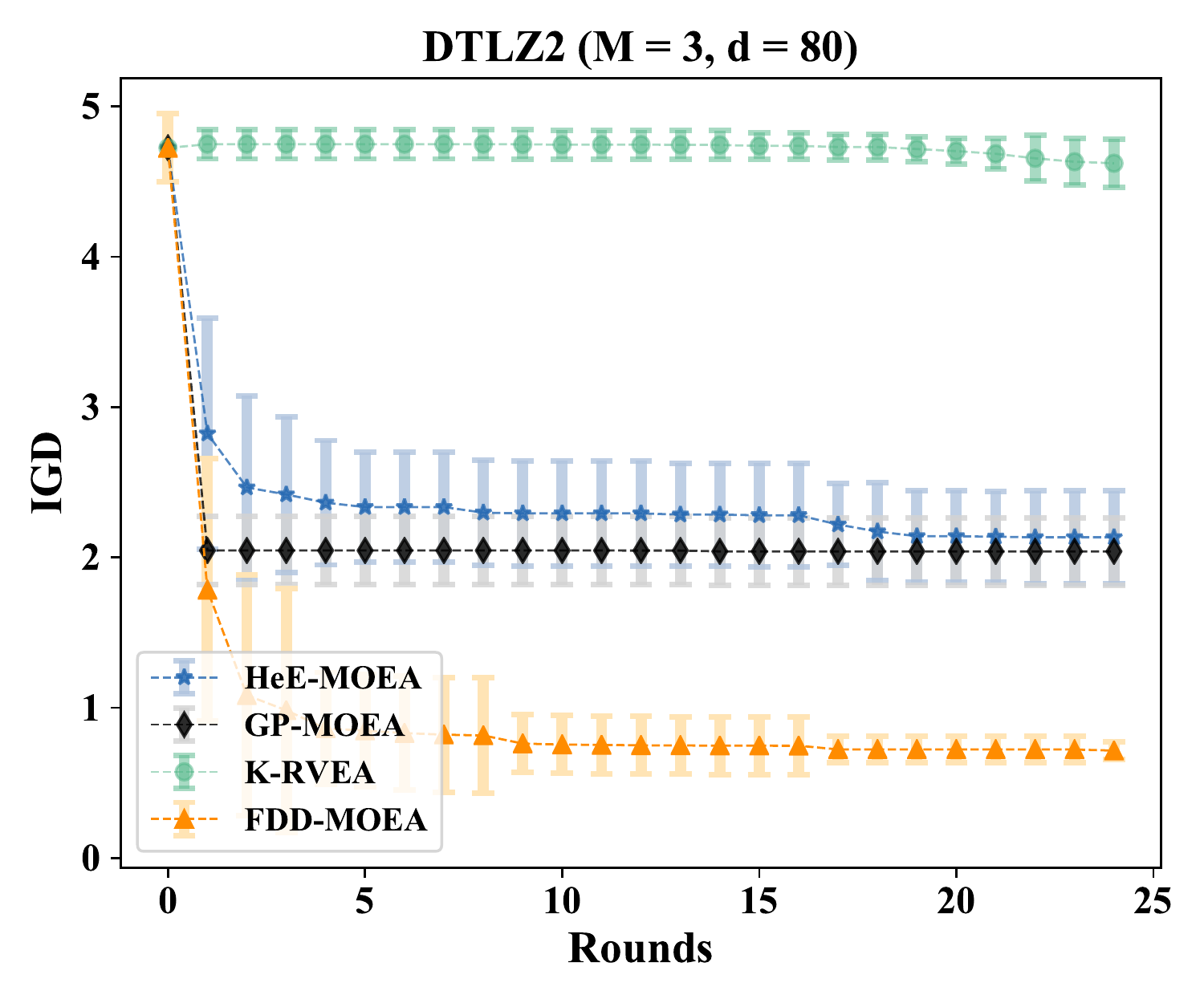} }}
    \caption{The IGD convergence profiles on DTLZ2.}
    \label{fig:DTLZ2}
\end{figure*}

\begin{figure*}
    \centering
    \subfigure[\centering DTLZ3, $M=3$, $d=10$]{{\includegraphics[width=0.48\textwidth]{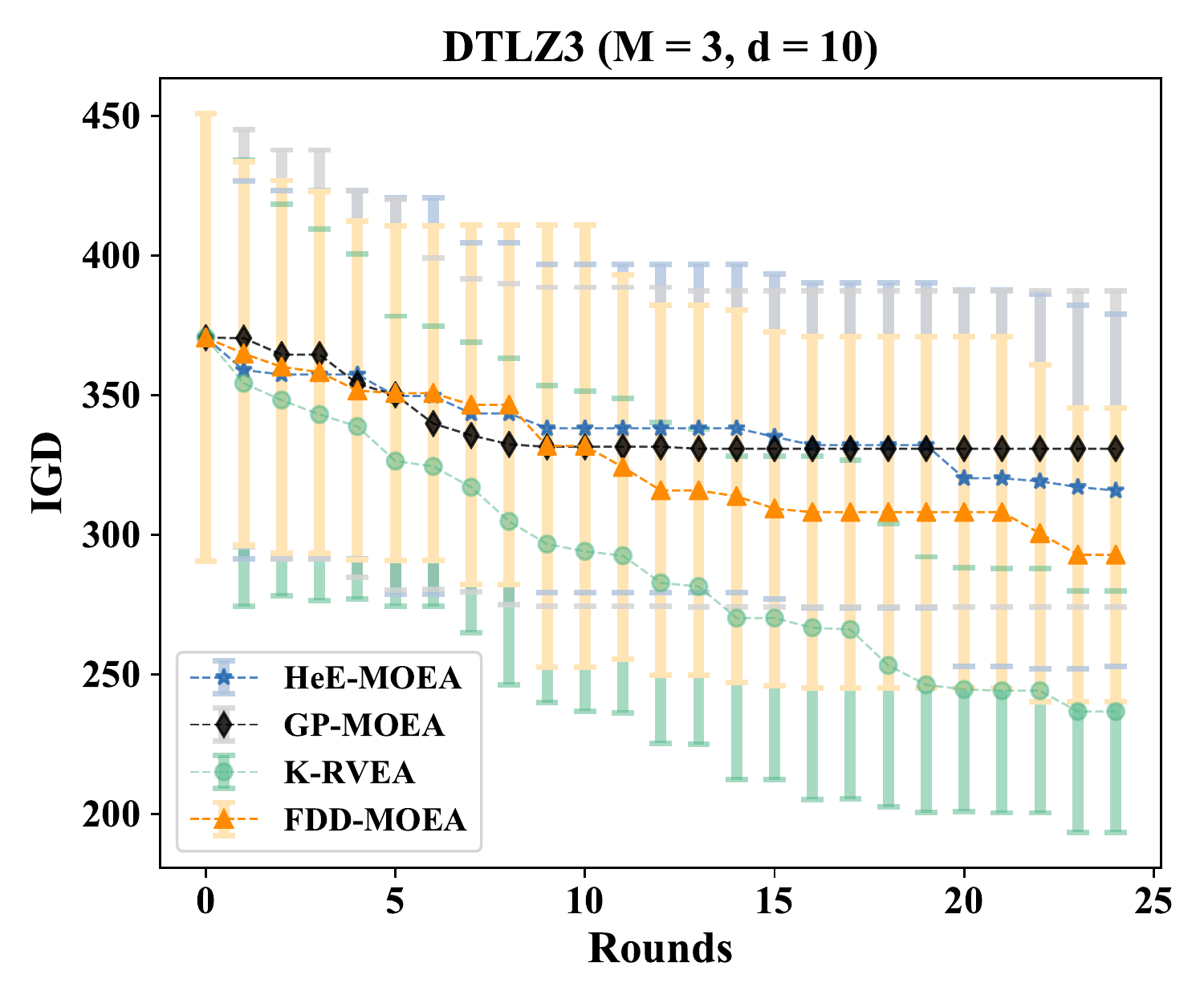} }}
    \subfigure[\centering DTLZ3, $M=3$, $d=80$]{{\includegraphics[width=0.48\textwidth]{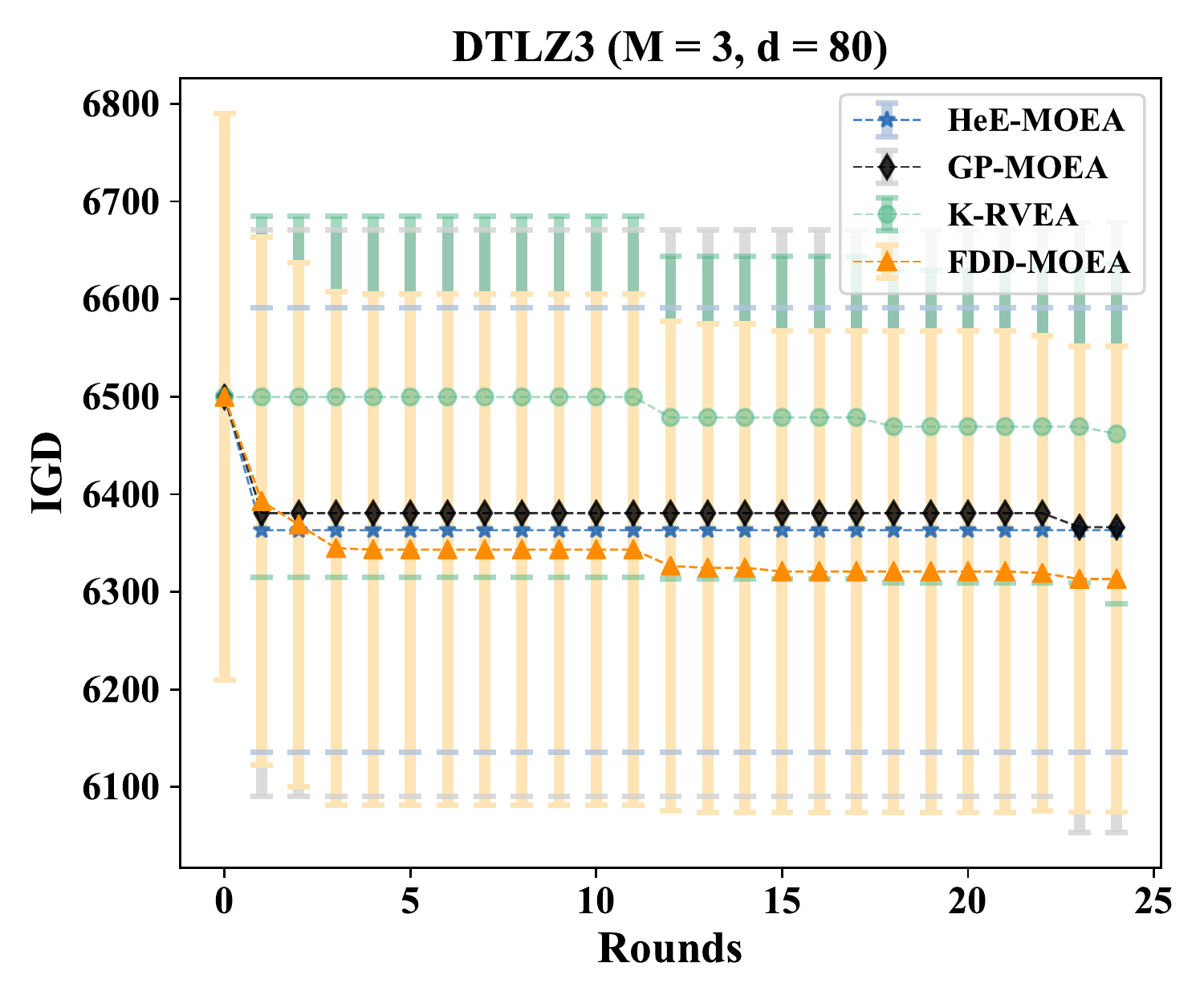} }}
    \caption{The IGD convergence profiles on DTLZ3.}
    \label{fig:DTLZ3}
\end{figure*}

\begin{figure*}
    \centering
    \subfigure[\centering DTLZ4, $M=3$, $d=10$]{{\includegraphics[width=0.48\textwidth]{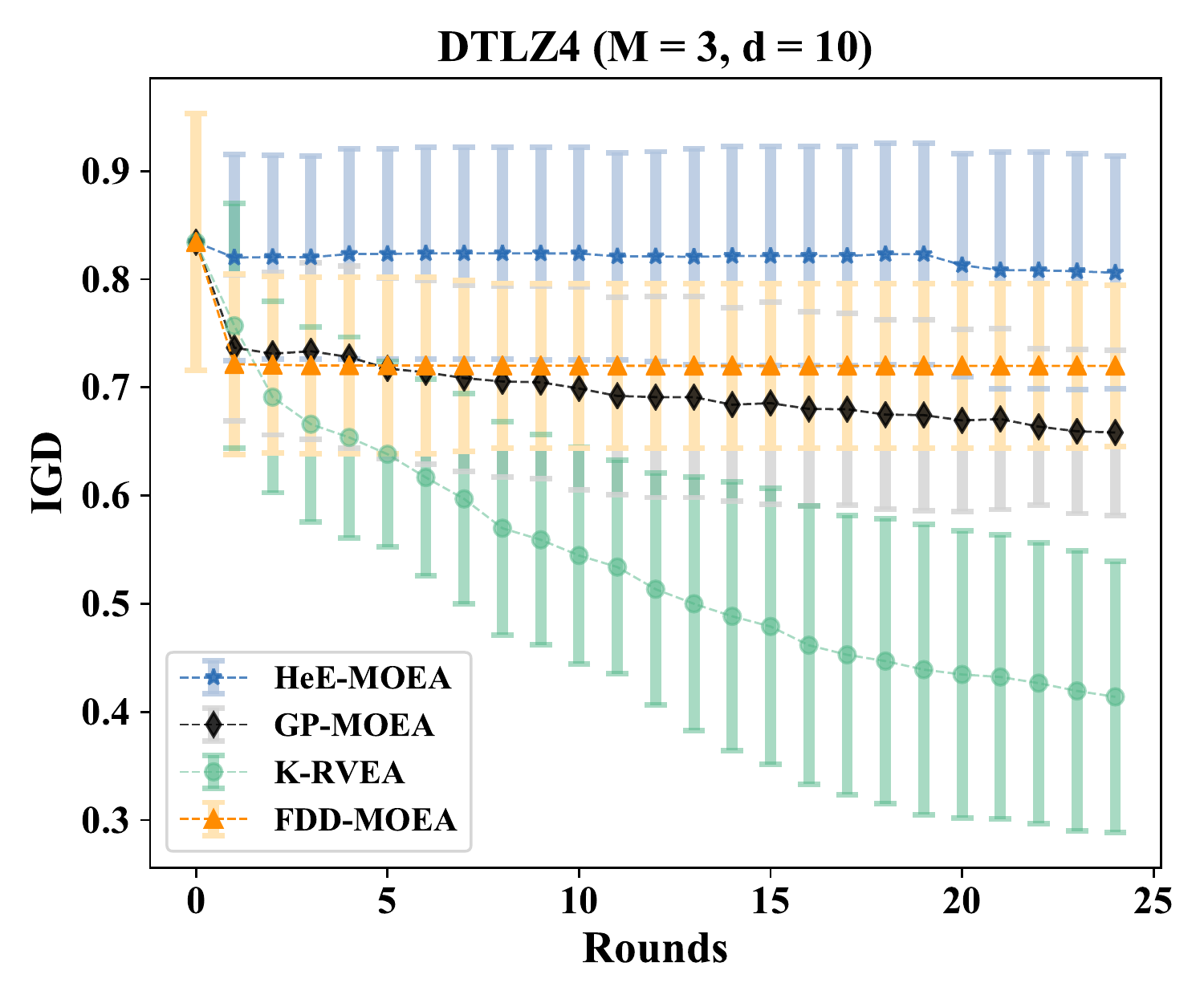} }}
    \subfigure[\centering DTLZ4, $M=3$, $d=80$]{{\includegraphics[width=0.48\textwidth]{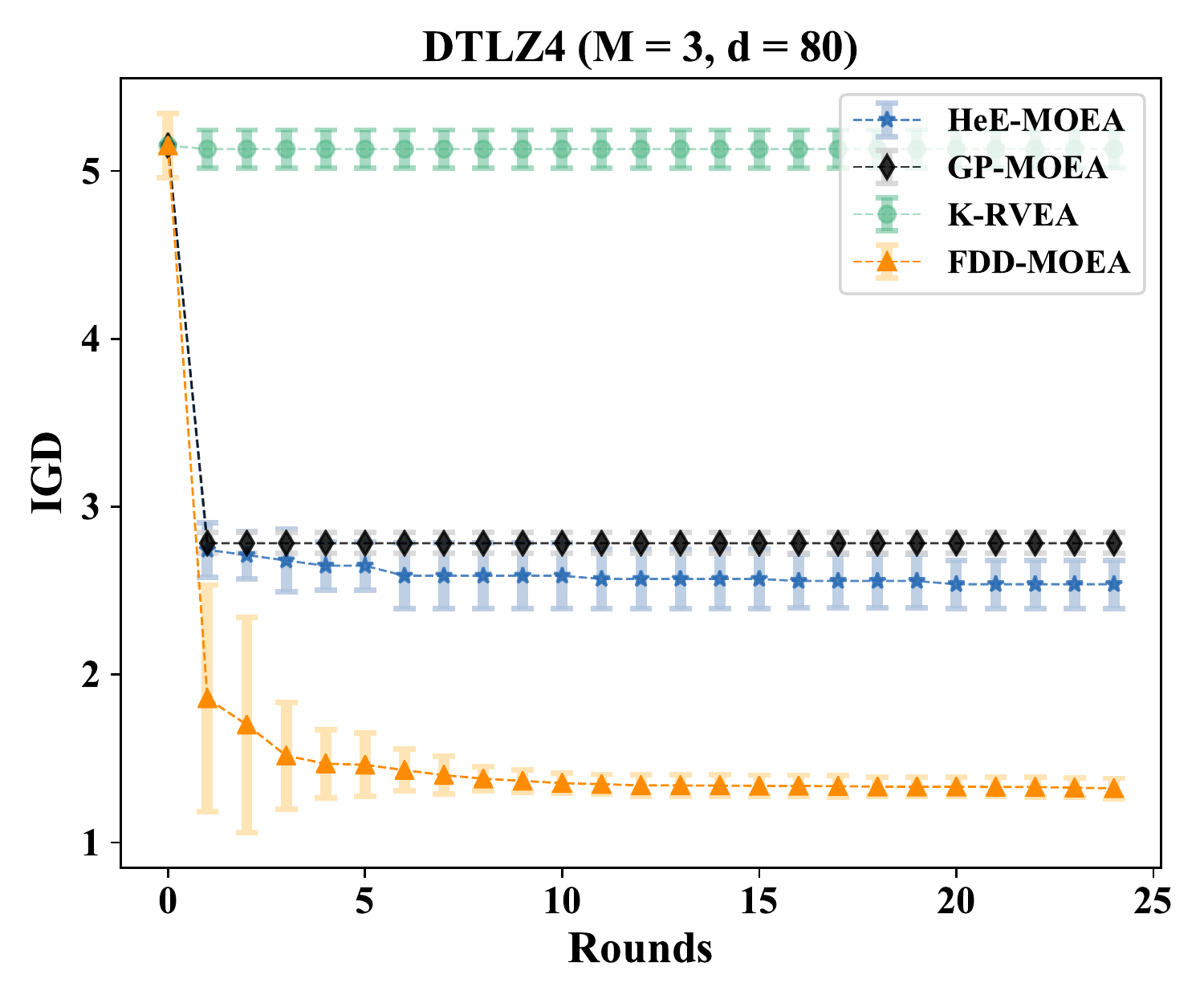} }}
    \caption{The IGD convergence profiles on DTLZ4.}
    \label{fig:DTLZ4}
\end{figure*}

\begin{figure*}
    \centering
    \subfigure[\centering DTLZ5, $M=3$, $d=10$]{{\includegraphics[width=0.48\textwidth]{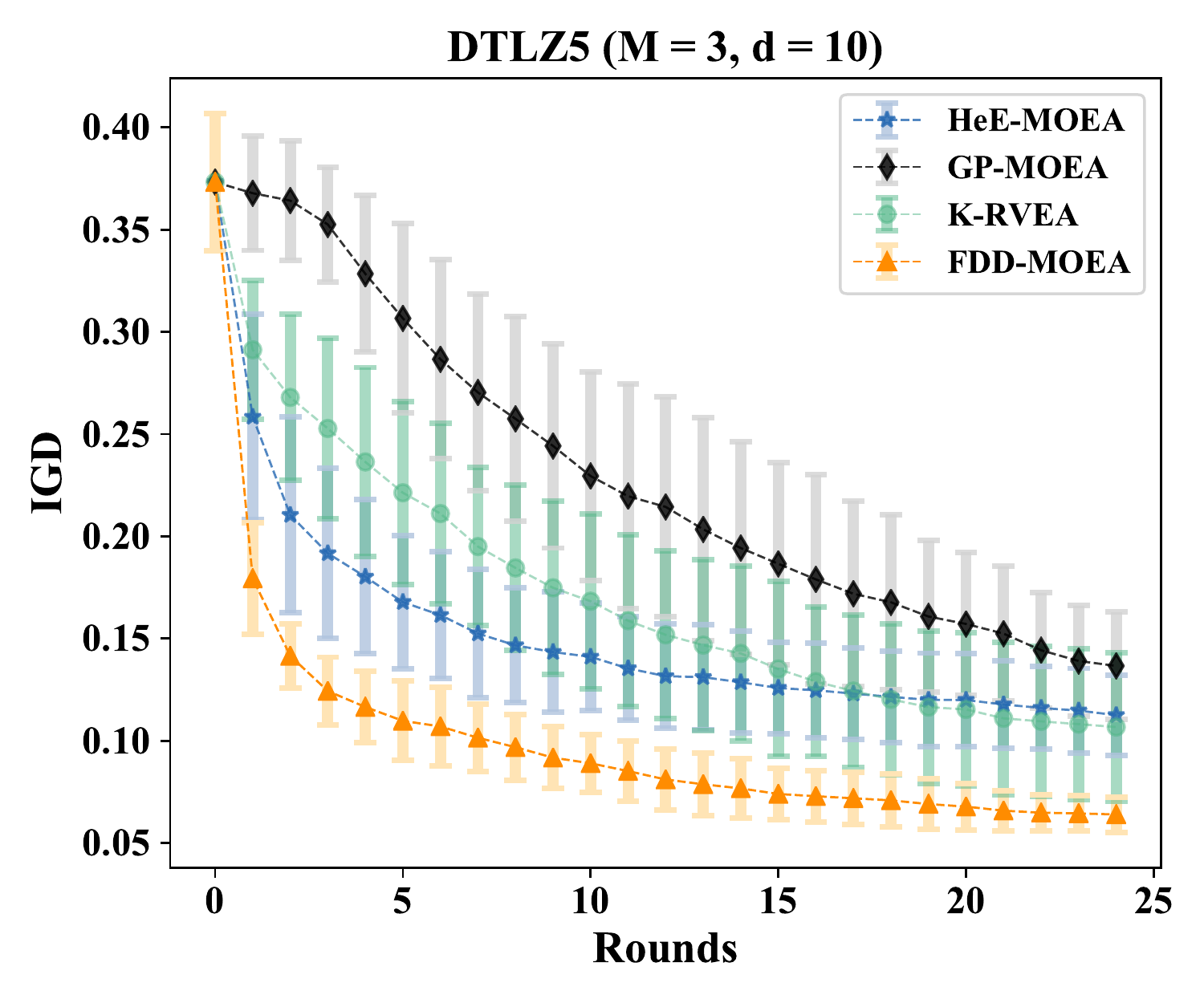} }}
    \subfigure[\centering DTLZ5, $M=3$, $d=80$]{{\includegraphics[width=0.48\textwidth]{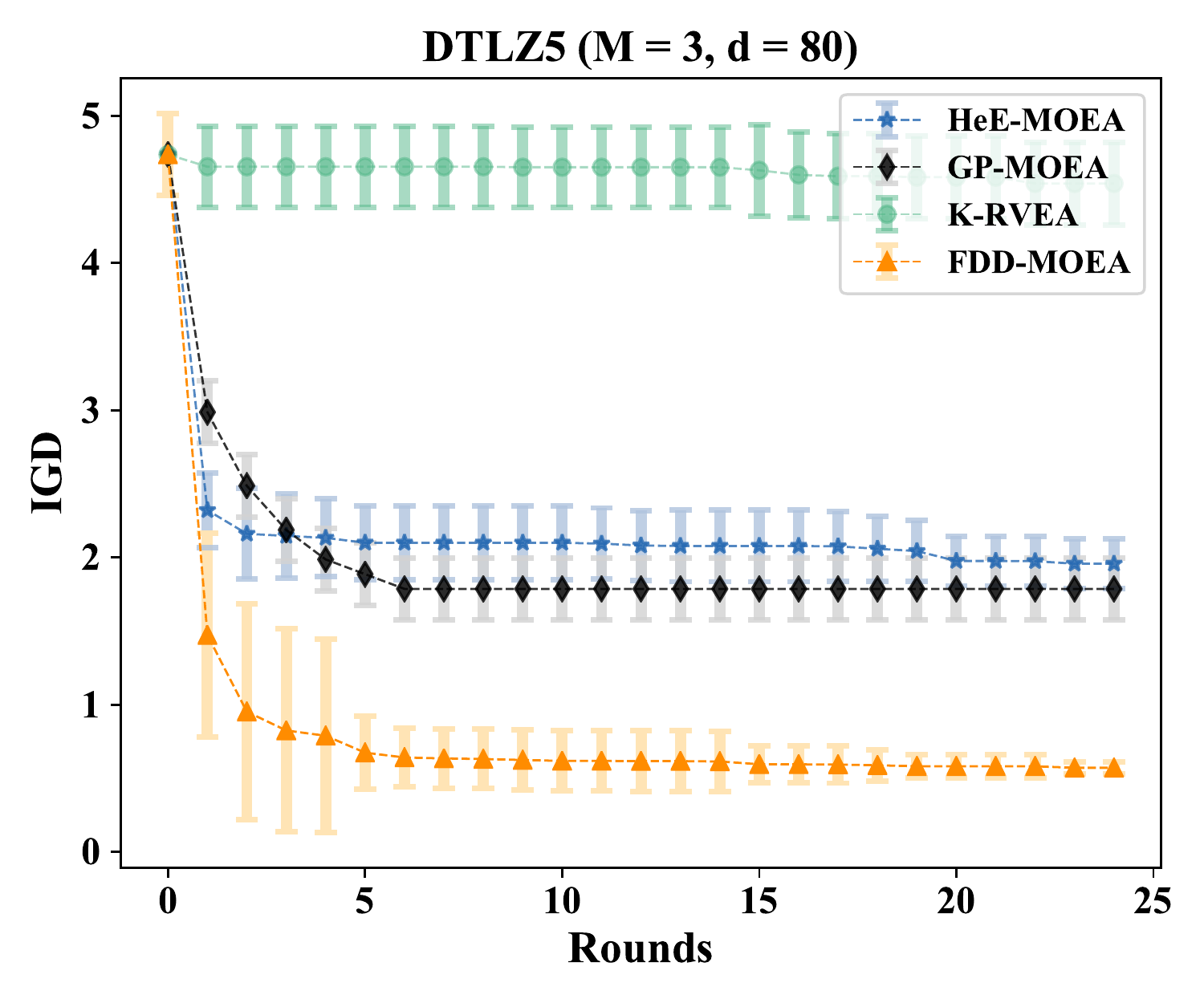} }}
    \caption{The IGD convergence profiles on DTLZ5.}
    \label{fig:DTLZ5}
\end{figure*}

\begin{figure*}
    \centering
    \subfigure[\centering DTLZ6, $M=3$, $d=10$]{{\includegraphics[width=0.48\textwidth]{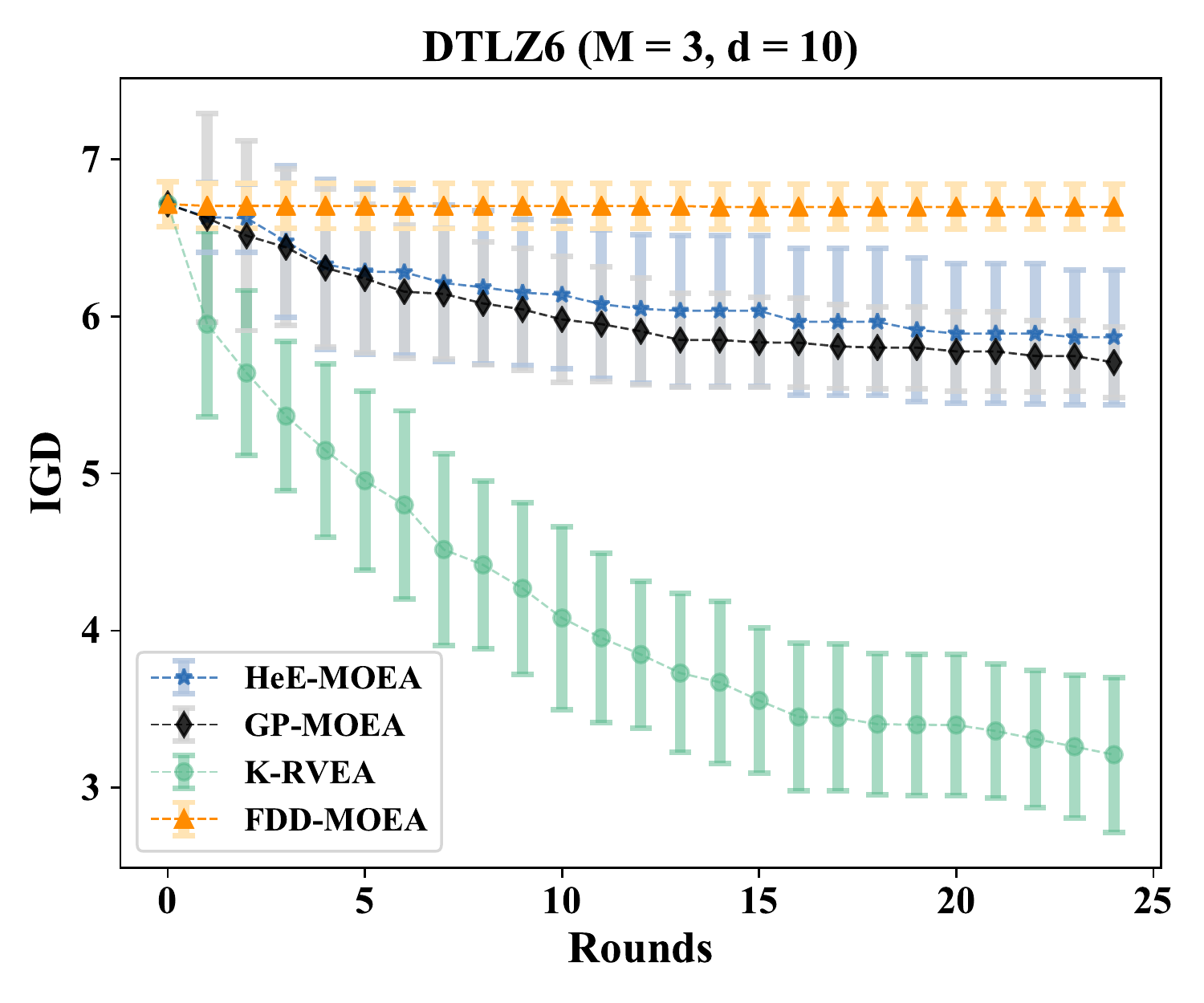} }}
    \subfigure[\centering DTLZ6, $M=3$, $d=80$]{{\includegraphics[width=0.48\textwidth]{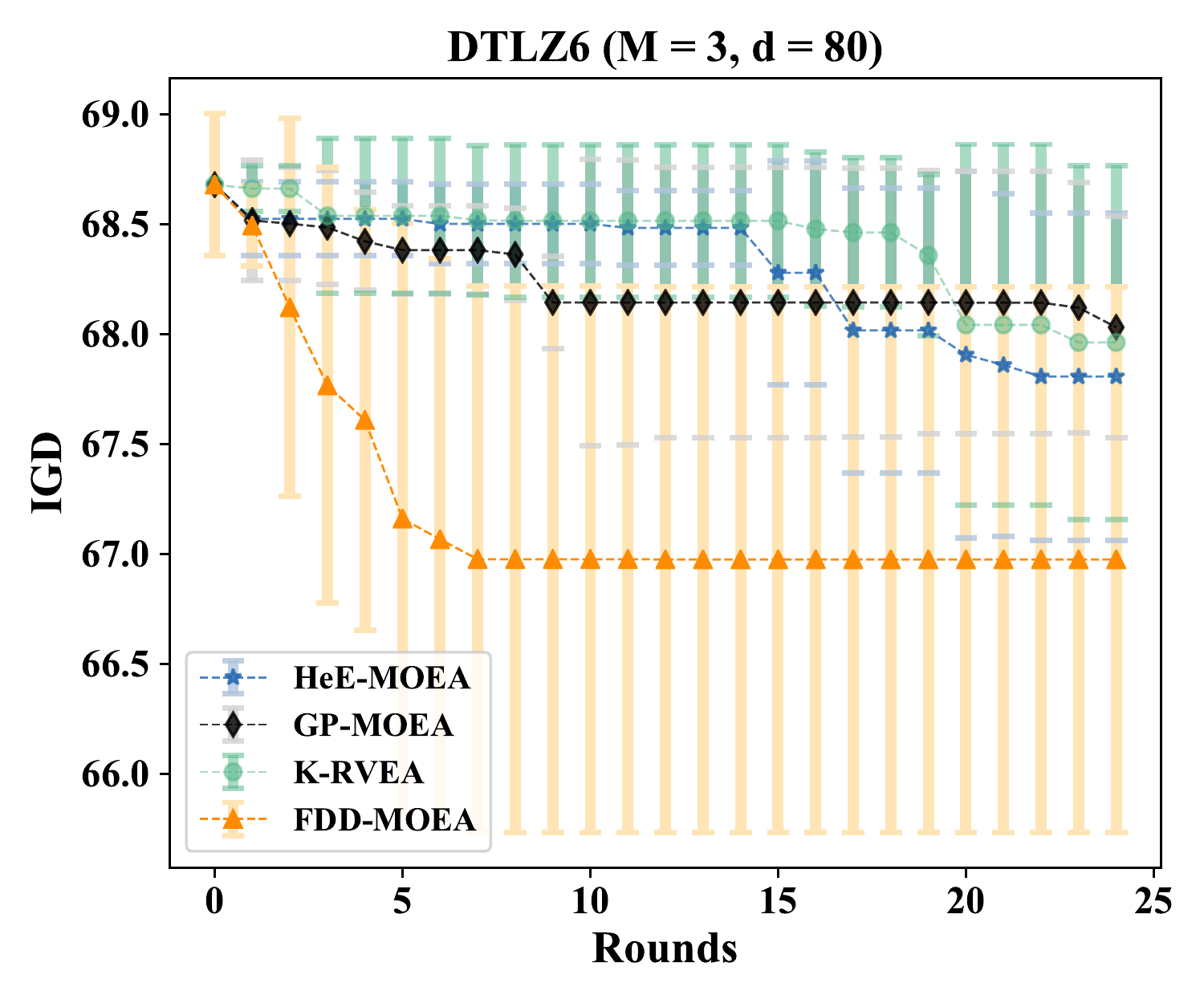} }}
    \caption{The IGD convergence profiles on DTLZ6.}
    \label{fig:DTLZ6}
\end{figure*}

\begin{figure*}
    \centering
    \subfigure[\centering DTLZ7, $M=3$, $d=10$]{{\includegraphics[width=0.48\textwidth]{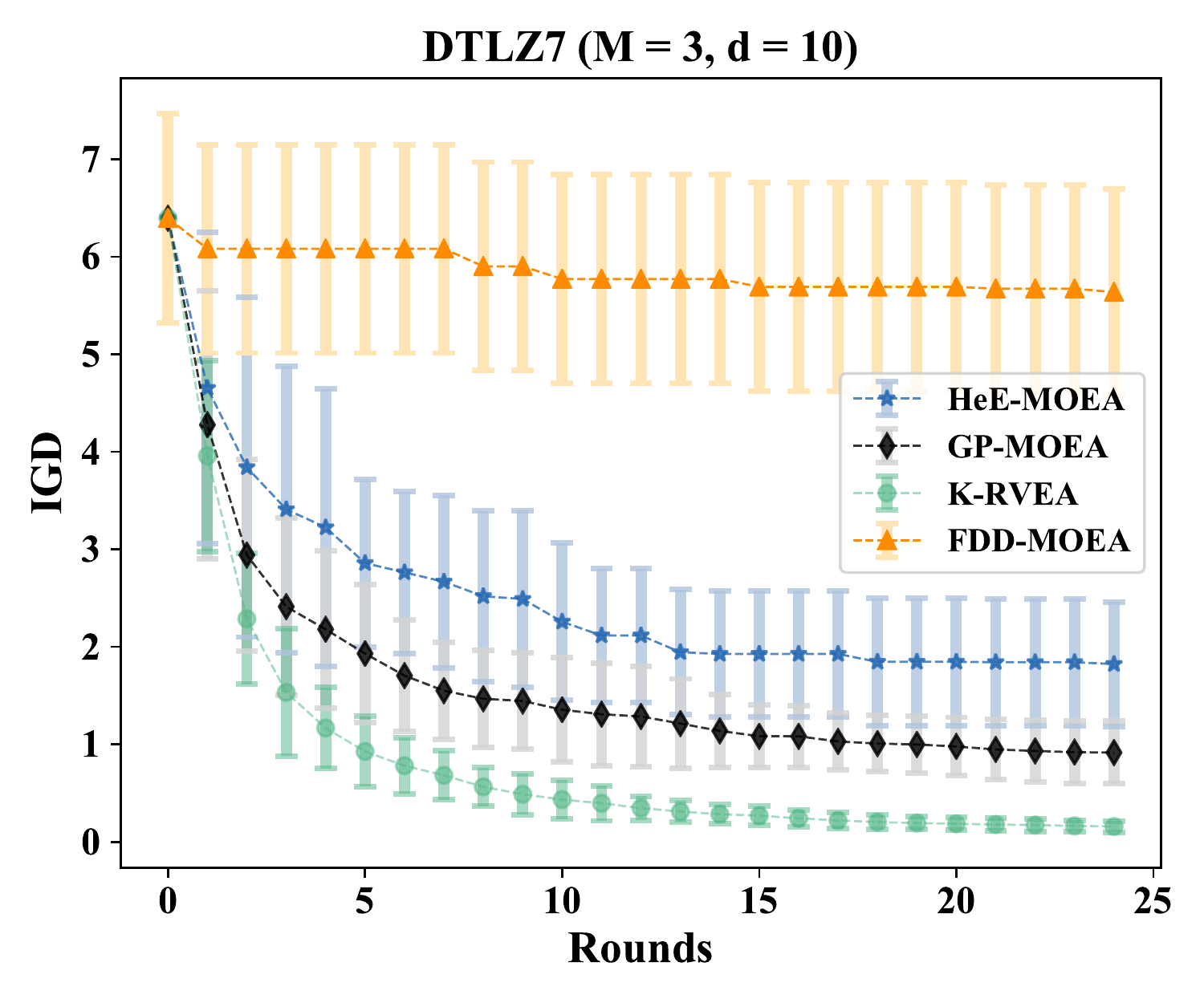} }}
    \subfigure[\centering DTLZ7, $M=3$, $d=80$]{{\includegraphics[width=0.48\textwidth]{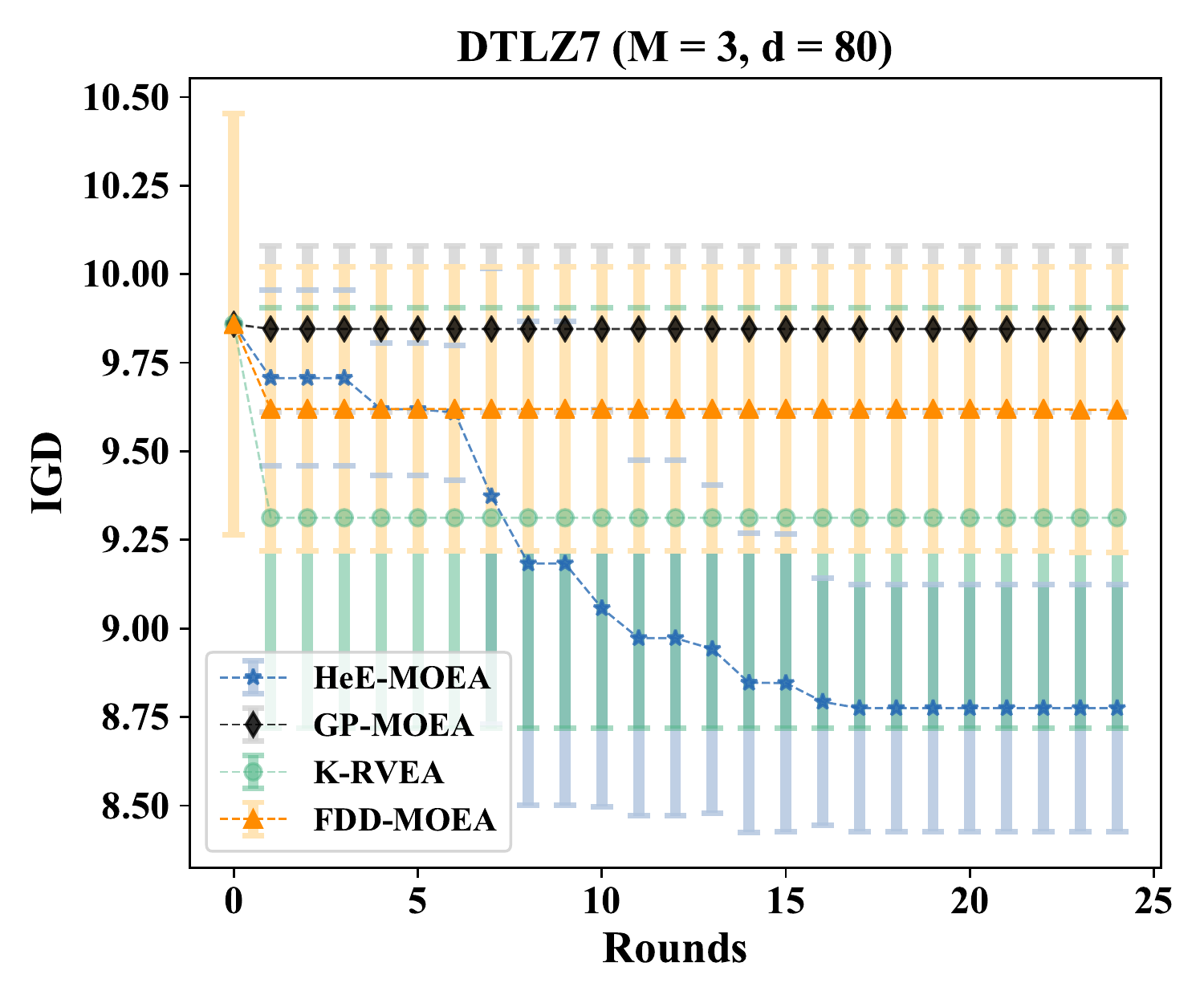} }\label{fig:DTLZ7_b}}
    \caption{The IGD convergence profiles on DTLZ7.}
    \label{fig:DTLZ7}
\end{figure*}

The convergence profiles of the compared algorithms on DTLZ1 - DTLZ7 when $d=10,80$ over the iterations are plotted in Figs. \ref{fig:DTLZ1}-\ref{fig:DTLZ7}. From these results, we find that the convergence properties of different algorithms strongly differ on different test instances. One general conclusion is that the performance advantage of FDD-MOEA over the compared algorithms will become stronger as the search dimension increases. This might be attributed to the use of the RBFNs in FDD-MOEA, which has been demonstrated powerful for high-dimensional systems in previous research. 




\subsubsection{Performance comparison on MaOPs}

To further examine the effectiveness of the proposed FDD-MOEA on distributed optimization of many-objective problems, we conduct experiments by setting the number of objectives $M=5,10,20$ for a fixed dimension $d=20$. In this case, all algorithms under comparison use RVEA as the baseline optimizer. Table \ref{Tab:M} presents the statistical results in terms of the IGD values. 

Overall, FDD-MOEA performs slightly better than HeE-MOEA and K-RVEA, and much better than GP-MOEA on the MaOP test instances. Compared with HeE-MOEA, FDD-MOEA performs better on 10 out of 21 test instances, equally well and worse on four and seven instances, respectively. Similar to the results presented above, the performance of FDD-MOEA is consistently competitive on all DTLZ1-DTLZ5 instances. FDD-MOEA performs particularly well on DTLZ6 when RVEA is adopted as the optimizer, in contrast to its performance when NSGA-II is used. On DTLZ7, FDD-MOEA is outperformed by HeE-MOEA. In addition, FDD-MOEA outperforms GP-MOEA on 15 out of 21 test instances, achieves similar results on three instances, and is underperformed on four instances.

In general, K-RVEA achieves the second best results on many-objective problems, and K-RVEA outperforms FDD-MOEA on the DTLZ2 and DTLZ7 test instances. Interestingly, when using RVEA as the optimizer, the performance of FDD-MOEA on DTLZ6 problems has a significant improvement and and it outperforms K-RVEA in all cases, which means that using RVEA as the optimizer makes FDD-MOEA more competitive when compared with K-RVEA.

\begin{table}
	\caption{Mean (std) of the IGD values obtained by FDD-MOEA, HeE-MOEA, GP-MOEA and K-RVEA for different numbers of objectives when $d=20$.}
	
	\label{Tab:M}
	\centering
	\renewcommand\arraystretch{1.8}
    \begin{tabular}{c|c|c|c|c|c}
    
    \hline
    Problems   & M     & FDD-MOEA        & HeE-MOEA         & GP-MOEA    & K-RVEA                 \\ \hline
    
    \multirow{3}{*}{DTLZ1}
    & 5   &\cellcolor{gray!50}212.97 (4.0e+01)  & 281.33 (2.4e+01) +  & 291.38 (2.7e+01) + & 236.16 (32.e+01) + \\ 
    & 10  & 130.24 (2.9e+01)  & 159.46 (2.2e+01) +  & 155.67 (2.5e+01) +  &\cellcolor{gray!50}110.32 (2.6e+01) - \\ 
    & 20  &\cellcolor{gray!50}0.4324 (7.9e-02)  & 0.4335 (7.8e-02) =  & 0.5084 (1.2e-01) = & 0.4398 (1.1e-01) = \\ \hline
    
    \multirow{3}{*}{DTLZ2}
    & 5   &\cellcolor{gray!50} 0.6704 (4.0e-02)  & 0.7940 (6.0e-02) +  & 0.9390 (5.9e-02) + & 0.7543 (5.7e-02) +  \\ 
    & 10  & 0.9413 (2.1e-02)  & 0.9172 (3.5e-02) -  & 1.0124 (4.3e-02) + &\cellcolor{gray!50}0.7958 (5.5e-02) -   \\ 
    & 20  & 1.0057 (1.4e-02)  & 1.0043 (1.4e-02) =  & 1.0103 (1.5e-02) = &\cellcolor{gray!50}0.9059 (3.5e-02) -  \\ \hline
    
    \multirow{3}{*}{DTLZ3} 
    & 5   &\cellcolor{gray!50}667.77 (9.6e+01)  & 939.50 (6.3e+01) +  & 909.28 (7.1e+01) + & 799.77 (7.7e+01) + \\ 
    & 10  &\cellcolor{gray!50}380.93 (7.9e+01)  & 538.91 (7.5e+01) +  & 548.25 (9.0e+01) + & 402.85 (6.0e+01) = \\ 
    & 20  & 1.4774 (2.1e-01)  & 1.4482 (2.2e-01) =  & 1.5485 (2.6e-01) = &\cellcolor{gray!50}1.3590 (1.6e-01) -  \\ \hline
    
    \multirow{3}{*}{DTLZ4} 
    & 5   & 1.1841 (5.4e-02)  & 1.1077 (1.9e-02) -  & 1.3370 (6.7e-02) + &\cellcolor{gray!50}1.0117 (1.1e-01) - \\ 
    & 10  &\cellcolor{gray!50}1.0113 (4.3e-02)  & 1.1466 (3.1e-02) =  & 1.2009 (4.2e-02) + & 1.0561 (7.3e-02) +  \\ 
    & 20  &\cellcolor{gray!50}0.7874 (9.1e-03)  & 0.8248 (1.1e-02) +  & 0.8276 (9.8e-03) + & 0.7943 (1.7e-02) +  \\ \hline
    
    \multirow{3}{*}{DTLZ5} 
    & 5   & 0.3709 (3.4e-02)  &\cellcolor{gray!50}0.2099 (4.0e-02) -  & 0.6635 (8.4e-02) + & 0.4276 (7.9e-02) + \\
    & 10  & 0.2654 (2.5e-02)  &\cellcolor{gray!50}0.1793 (3.2e-02) -  & 0.4124 (5.6e-02) + & 0.2534 (4.4e-02) =  \\ 
    & 20  &\cellcolor{gray!50}0.0109 (1.1e-03)  & 0.0118 (1.1e-03) +  & 0.0126 (1.3e-03) + & 0.0118 (1.2e-03) +  \\ \hline
    
    \multirow{3}{*}{DTLZ6} 
    & 5   &\cellcolor{gray!50}8.4899 (1.2e+00)  & 13.100 (5.4e-01) +  & 12.627 (3.7e-01) + & 9.9443 (7.6e-01) + \\
    & 10  &\cellcolor{gray!50}5.8207 (7.8e-01)  & 8.8257 (2.8e-01) +  & 8.5187 (3.1e-01) + & 7.4196 (5.1e-01) +  \\
    & 20  & 0.2301 (1.1e-01)  & 0.5749 (1.2e-01) +  & 0.4608 (9.2e-02) + &\cellcolor{gray!50}0.1929 (1.1e-01) = \\ \hline
    
    \multirow{3}{*}{DTLZ7} 
    & 5   & 12.584 (1.6e+00)  & 4.6368 (8.1e-01) -  &\cellcolor{gray!50}1.2730 (2.6e-01) - & 1.2727 (4.6e-01) -  \\ 
    & 10  & 22.764 (2.7e+00)  & 5.9195 (1.5e+00) -  & 4.7156 (6.4e-01) - &\cellcolor{gray!50}2.3166 (1.3e-01) - \\ 
    & 20  & 4.7349 (1.1e+00)  & 3.6157 (6.3e-01) -  & 2.9332 (8.1e-02) - &\cellcolor{gray!50}2.6058 (4.9e-01) - \\ \hline
    
    \multicolumn{2}{c|}{+/-/=}  & -  &  10/7/4  & 15/3/3 & 9/8/4   \\ \hline
    
    \end{tabular}
\end{table}

\subsubsection{Influence of $\lambda$}
The participation ratio, $\lambda$, is an important parameter in federated learning systems. In federated data-driven optimization, a low participation rate will lead to a decrease in the amount of training data on each client. To examine the influence of the participation ratio $\lambda$ on the performance of FDD-MOEA, empirical experiments are conducted by varying the values of $\lambda$, while the number of decision variables is fixed to $d=30$ and RVEA is adopted as the optimizer. 

\begin{figure*}[htbp]
    \centering
    \includegraphics[width=0.95\textwidth]{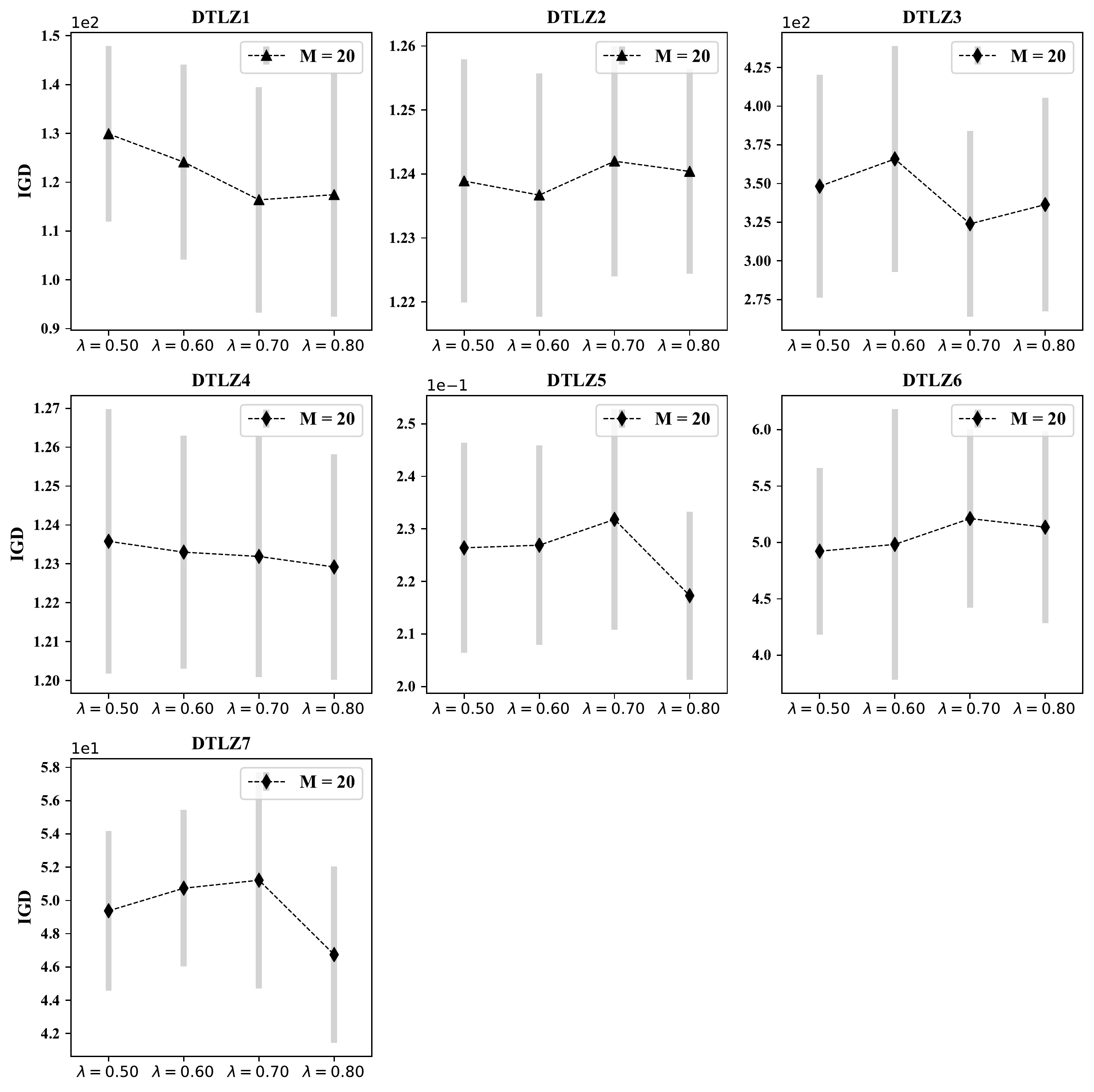}
    \caption{Converged IGD values when $M=20, d=30$ with different $\lambda$.}
    \label{fig:lambda}   
\end{figure*}

As listed in Table \ref{Tab:lambda}, the results indicate that FDD-MOEA is relatively insensitive to the participation ratio and in general, a larger participation ratio contributes to better performance, which is expected. Furthermore, we also investigate the influence of $\lambda$ on the performance of FDD-MOEA DTLZ1-7 problems in terms of IGD for $M=20, d=30$. The performance changes of FDD-MOEA over different $\lambda$ values are presented in Fig. \ref{fig:lambda}. The experimental results show that higher participation ratios will lead to an improvement in the performance of FDD-MOEA, mainly because the total amount of selected solutions is limited, and the amount of local training data on each client may significantly decrease with a lower participation ratio $\lambda$.

\begin{table}
	\caption{Mean (std) of the IGD values obtained by using different $\lambda$ for different numbers of objectives when $d=30$.}
	
	\label{Tab:lambda}
	\centering
	\renewcommand\arraystretch{1.8}
    \begin{tabular}{c|c|c|c|c|c}
    
    \hline
    Problems   & M     & $\lambda=0.50$        & $\lambda=0.60$         & $\lambda=0.70$    & $\lambda=0.80$                 \\ \hline
    
    \multirow{4}{*}{DTLZ1}
    & 3   & 475.72 (6.4e+01)  & 496.86 (7.1e+01)   & 471.51 (7.5e+01)  &\cellcolor{gray!50}457.47 (6.0e+01)   \\ 
    & 5   & 398.62 (6.3e+01)  & 400.85 (6.2e+01)   & 389.34 (5.7e+01)  &\cellcolor{gray!50}380.88 (6.1e+01)  \\ 
    & 10  & 306.73 (2.6e+01)  & 314.13 (3.6e+01)   &\cellcolor{gray!50}299.45 (4.9e+01)  & 303.69 (4.6e+01)  \\ 
    & 20  & 129.90 (1.8e+01)  & 124.11 (2.0e+01)   &\cellcolor{gray!50}116.39 (2.3e+01)  & 117.44 (2.5e+01)  \\ \hline
    
    \multirow{4}{*}{DTLZ2}
    & 3   & 0.6341 (3.2e-02)  &\cellcolor{gray!50}0.6197 (5.6e-02)   & 0.6616 (6.6e-02)  & 0.6439 (5.2e-02)  \\
    & 5   &\cellcolor{gray!50}0.9431 (5.2e-02)  & 0.9586 (4.3e-02)   & 0.9878 (5.5e-02)  & 0.9665 (6.3e-02)  \\ 
    & 10  & 1.1574 (3.2e-02)  & 1.1349 (4.3e-02)   & 1.1651 (4.3e-02)  &\cellcolor{gray!50}1.1255 (5.2e-02)  \\ 
    & 20  & 1.2389 (1.9e-02)  &\cellcolor{gray!50}1.2367 (1.9e-02)   & 1.2420 (1.8e-02)  & 1.2404 (1.6e-02)  \\ 
    \hline
    
    \multirow{4}{*}{DTLZ3} 
    & 3   & 1135.4 (2.1e+02)  & 1177.3 (1.9e+02)   & 1163.9 (2.3e+02)  &\cellcolor{gray!50}1083.4 (1.4e+02)  \\
    & 5   & 1173.5 (1.7e+02)  & 1118.1 (1.1e+02)   &\cellcolor{gray!50}1055.4 (1.9e+02)  & 1066.2 (1.5e+02)  \\ 
    & 10  & 775.74 (1.4e+02)  & 790.38 (1.4e+02)   & 811.34 (1.4e+02)  &\cellcolor{gray!50}771.46 (1.1e+02)  \\ 
    & 20  & 348.21 (7.2e+01)  & 365.86 (7.3e+01)   &\cellcolor{gray!50}323.94 (6.0e+01)  & 336.40 (6.9e+01)  \\ 
    \hline
    
    \multirow{4}{*}{DTLZ4} 
    & 3   & 1.2373 (9.9e-02)  & 1.2001 (8.7e-02)   & 1.2321 (7.3e-02)  &\cellcolor{gray!50}1.1815 (6.7e-02)  \\
    & 5   & 1.4359 (4.7e-02)  & 1.4242 (6.5e-02)   &\cellcolor{gray!50}1.4235 (6.7e-02)  & 1.4427 (6.3e-02)  \\ 
    & 10  &\cellcolor{gray!50}1.4256 (4.7e-02)  & 1.4541 (6.4e-02)   & 1.4396 (6.4e-02)  & 1.4522 (4.6e-02)  \\ 
    & 20  & 1.2358 (3.4e-02)  & 1.2330 (3.0e-02)   & 1.2319 (3.1e-02)  &\cellcolor{gray!50}1.2292 (2.9e-02)  \\
    \hline
    
    \multirow{4}{*}{DTLZ5} 
    & 3   &\cellcolor{gray!50}0.5427 (5.6e-02)  & 0.5940 (6.7e-02)   & 0.5785 (6.8e-02)  & 0.5651 (7.2e-02)  \\
    & 5   & 0.7087 (7.8e-02)  & 0.7082 (7.5e-02)   & 0.6974 (6.0e-02)  &\cellcolor{gray!50}0.6608 (6.4e-02)  \\ 
    & 10  & 0.5259 (4.8e-02)  & 0.5091 (4.7e-02)   & 0.5369 (4.3e-02)  &\cellcolor{gray!50}0.5052 (5.7e-02)  \\ 
    & 20  & 0.2264 (2.0e-02)  & 0.2269 (1.9e-02)   & 0.2318 (2.1e-02)  &\cellcolor{gray!50}0.2173 (1.6e-02)  \\ 
    \hline
    
    \multirow{4}{*}{DTLZ6} 
    & 3   & 15.439 (1.2e+00)  & 15.305 (1.4e+00)   & 15.439 (1.4e+00)  &\cellcolor{gray!50}14.819 (1.4e+00)  \\
    & 5   & 14.102 (1.5e+00)  & 13.438 (1.3e+00)   & 14.057 (1.5e+00)  &\cellcolor{gray!50}12.985 (1.6e+00)  \\ 
    & 10  & 11.013 (1.3e+00)  & 11.130 (1.1e+00)   &\cellcolor{gray!50}10.910 (1.2e+00)  & 11.079 (1.3e+00)  \\ 
    & 20  &\cellcolor{gray!50}4.9225 (7.4e-01)  & 4.9826 (1.2e+00)   & 5.2115 (7.9e-01)  & 5.1356 (8.5e-01)  \\
    \hline
    
    \multirow{4}{*}{DTLZ7} 
    & 3   & 8.4360 (9.8e-01)  & 8.4062 (7.7e-01)   &\cellcolor{gray!50}7.8610 (1.1e+00)  & 8.0583 (8.1e-01)  \\
    & 5   & 14.426 (1.2e+00)  & 14.386 (1.5e+00)   &\cellcolor{gray!50}14.275 (1.2e+00)  & 14.396 (2.2e+00)  \\ 
    & 10  & 28.383 (2.1e+00)  & 28.492 (2.0e+00)   & 27.903 (2.0e+00)  &\cellcolor{gray!50}27.884 (3.0e+00)  \\ 
    & 20  & 49.375 (4.8e+00)  & 50.736 (4.7e+00)   & 51.217 (6.5e+00)  &\cellcolor{gray!50}46.742 (5.3e+00)  \\
    \hline
    
    
    \end{tabular}
\end{table}

\subsubsection{Influence of $p_f$}
\begin{figure*}[htbp]
    \centering
    \includegraphics[width=0.95\textwidth]{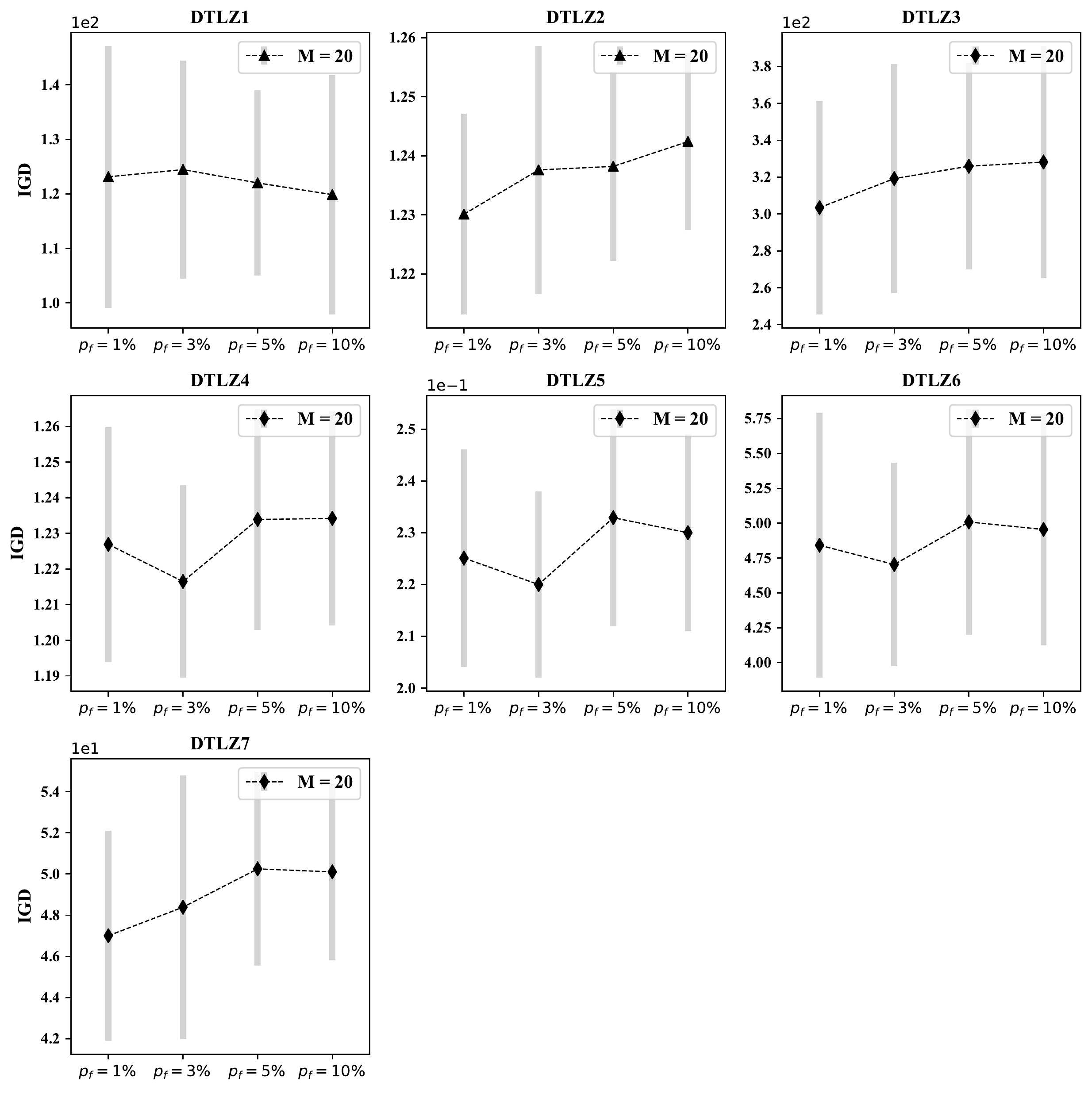}
    \caption{Converged IGD values when $M=20, d=30$ with different $p_f$.}
    \label{fig:pf}   
\end{figure*}

The aim of federated data-driven EAs is to solve of distributed optimization problems without collecting private data stored on local clients \cite{xu2021federated} and hence, communications between clients and the server when exchanging surrogate information are unavoidable. However, there are many potential reasons that may cause communication failures or packet losses, such as network congestion or when communication network is not available. As mentioned in Section \ref{sec:framework}, we assume that each client has a failure probability of $p_f$ when communicating with the server, which means that the $m$ solutions obtained in this iteration cannot be passed to the client. In this part, we study the impact of $p_f$ on the performance of FDD-MOEA, and the experimental results with $p_f=1 \%,3 \%,5 \%$ and $10 \%$ are summarized in Table \ref{Tab:pf}, and the converged IGD values of the compared algorithms are plotted in Fig. \ref{fig:pf} when $M=20$.

Specifically, the dimension of the decision space $d$ and the participation ratio are fixed to 30 and 0.90, respectively, and RVEA is set as the optimizer. As we can see, when $p_f$ is set to 1 \% and 3 \%, the performance of FDD-MOEA does not change much and becomes slightly when $p_f$ is increased to 5\% or 10\%. This is also expected since the larger $p_f$ is, the less data will be available for model management on each client. 

\begin{table}
	\caption{Mean (std) of the IGD values obtained by using different $p_f$ for different numbers of objectives when $d=30$.}
	
	\label{Tab:pf}
	\centering
	\renewcommand\arraystretch{1.8}
    \begin{tabular}{c|c|c|c|c|c}
    
    \hline
    Problems   & M     & $p_f=1\%$        & $p_f=3\%$         & $p_f=5\%$    & $p_f=10\%$                 \\ \hline
    
    \multirow{4}{*}{DTLZ1}
    & 3   &\cellcolor{gray!50}425.28 (5.9e+01)  & 428.37 (8.1e+01)   & 443.01 (9.1e+01)  & 456.45 (6.6e+01)  \\ 
    & 5   & 401.79 (5.3e+01)  &\cellcolor{gray!50}370.45 (5.7e+01)   & 381.59 (5.4e+01)  & 406.36 (6.4e+01)  \\ 
    & 10  &\cellcolor{gray!50}289.57 (5.1e+01)  & 293.33 (5.7e+01)   & 294.34 (4.9e+01)  & 295.57 (5.2e+01)  \\ 
    & 20  & 123.13 (2.4e+01)  & 124.45 (2.0e+01)   & 122.01 (1.7e+01)  &\cellcolor{gray!50}119.86 (2.2e+01)  \\ \hline
    
    \multirow{4}{*}{DTLZ2}
    & 3   &\cellcolor{gray!50}0.6255 (4.8e-02)  & 0.6432 (5.9e-02)   & 0.6295 (5.4e-02)  & 0.6526 (4.6e-02)  \\
    & 5   & 0.9737 (4.5e-02)  & 0.9676 (5.2e-02)   &\cellcolor{gray!50}0.9596 (5.8e-02)  & 0.9984 (5.4e-02)  \\ 
    & 10  & 1.1639 (5.1e-02)  &\cellcolor{gray!50}1.1481 (3.4e-02)   & 1.1628 (5.2e-02)  & 1.1549 (4.9e-02)  \\ 
    & 20  &\cellcolor{gray!50}1.2301 (1.7e-02)  & 1.2376 (2.1e-02)   & 1.2382 (1.6e-02)  & 1.2424 (1.5e-02)  \\ \hline
    
    \multirow{4}{*}{DTLZ3} 
    & 3   & 1055.2 (1.1e+02)  & 1057.7 (1.6e+02)   &\cellcolor{gray!50}1014.7 (1.4e+02)  & 1108.5 (1.6e+02)  \\
    & 5   & 1036.4 (1.7e+02)  & 1001.7 (2.3e+02)   & 1066.3 (1.6e+02)  &\cellcolor{gray!50}1025.0 (2.1e+02)  \\ 
    & 10  & 710.35 (1.5e+02)  &\cellcolor{gray!50}689.86 (1.2e+02)   & 811.14 (1.4e+02)  & 793.48 (1.6e+02)  \\ 
    & 20  &\cellcolor{gray!50}303.42 (5.8e+01)  & 319.20 (6.2e+01)   & 325.92 (5.6e+01)  & 328.17 (6.3e+01)  \\ \hline
    
    \multirow{4}{*}{DTLZ4} 
    & 3   & 1.2583 (8.7e-02)  & 1.2479 (7.9e-02)   & 1.2221 (7.1e-02)  &\cellcolor{gray!50}1.2166 (9.9e-02)  \\
    & 5   & 1.4474 (7.0e-02)  &\cellcolor{gray!50}1.4226 (6.2e-02)   & 1.4283 (4.9e-02)  & 1.4276 (7.5e-02)  \\ 
    & 10  &\cellcolor{gray!50}1.4347 (4.4e-02)  & 1.4493 (2.7e-02)   & 1.4681 (4.8e-02)  & 1.4540 (5.2e-02)  \\ 
    & 20  & 1.2269 (3.3e-02)  &\cellcolor{gray!50}1.2165 (2.7e-02)   & 1.2339 (3.1e-02)  & 1.2342 (3.0e-02)  \\
    \hline
    
    \multirow{4}{*}{DTLZ5} 
    & 3   & 0.5753 (6.2e-02)  &\cellcolor{gray!50}0.5554 (5.8e-02)   & 0.5708 (4.5e-02)  & 0.5725 (6.5e-02)  \\
    & 5   &\cellcolor{gray!50}0.6717 (8.8e-02)  & 0.7073 (3.9e-02)   & 0.7044 (6.4e-02)  & 0.6980 (6.3e-02)  \\ 
    & 10  &\cellcolor{gray!50}0.5118 (7.5e-02)  & 0.5363 (5.5e-02)   & 0.5153 (6.2e-02)  & 0.5217 (6.1e-02)  \\ 
    & 20  & 0.2251 (2.1e-02)  &\cellcolor{gray!50}0.2200 (1.8e-02)   & 0.2329 (2.1e-02)  & 0.2300 (1.9e-02)  \\ 
    \hline
    
    \multirow{4}{*}{DTLZ6} 
    & 3   & 14.487 (1.5e+00)  & 14.060 (2.0e+00)   & 14.562 (1.5e+00)  &\cellcolor{gray!50}13.952 (1.3e+00)  \\
    & 5   &\cellcolor{gray!50}12.484 (2.0e+00)  & 13.263 (2.0e+00)   & 13.457 (1.4e+00)  & 12.686 (1.5e+00)  \\ 
    & 10  &\cellcolor{gray!50}9.8617 (1.6e+00)  & 10.235 (1.3e+00)   & 10.400 (9.1e-01)  & 10.425 (1.5e+00)  \\ 
    & 20  & 4.8421 (9.5e-01)  &\cellcolor{gray!50}4.7037 (7.3e-01)   & 5.0091 (8.1e-01)  & 4.9551 (8.3e-01)  \\
    \hline
    
    \multirow{4}{*}{DTLZ7} 
    & 3   &\cellcolor{gray!50}8.0467 (1.0e+00)  & 8.3347 (8.5e-01)   & 8.2695 (8.5e-01)  & 8.2858 (8.6e-01)  \\
    & 5   &\cellcolor{gray!50}13.971 (2.0e+00)  & 14.128 (1.4e+00)   & 14.413 (1.4e+00)  & 14.118 (1.9e+00)  \\ 
    & 10  & 28.991 (1.7e+00)  & 28.430 (2.2e+00)   &\cellcolor{gray!50}27.823 (2.7e+00)  & 28.451 (1.7e+00)  \\ 
    & 20  &\cellcolor{gray!50}47.002 (5.1e+00)  & 48.388 (6.4e+00)   & 50.248 (4.7e+00)  & 50.102 (4.3e+00)  \\
    \hline
    
    
    \end{tabular}
\end{table}

\section{Conclusions and future work}
\label{sec:conclusion}
It may become unrealistic to assume that all data can be made available for centralized training of surrogates in data-driven evolutionary optimization. To remove this strong assumption, we consider the challenging situation in which data are distributed on different local clients and propose a federated data-driven multi/many-objective evolutionary optimization algorithm in this work. By leveraging federated learning, a global surrogate is trained collaboratively for fitness approximations, a sorted averaging algorithm is used to aggregate the local models, and the uploaded local surrogates are utilized to provide the uncertainty of solutions. In addition, a modified acquisition function combining local and global information is proposed for surrogate management. Empirical results indicate that the proposed algorithm achieves comparable results on 49 benchmark problems with the state-of-the-art of centralized data-driven evolutionary algorithms. 

However, the proposed algorithm does not perform well on the problems with disconnected Pareto fronts, in which case the local model may strongly differ from each other, causing considerable model divergence. Hence, in the future, we will investigate more robust modelling techniques and new acquisition functions for handling strongly non-iid data for data-driven optimization.

\begin{acknowledgements}
This work was supported by National Natural Science Foundation of China (Basic Science Center Program: 61988101), International (Regional) Cooperation and Exchange Project(61720106008), National Natural Science Fund for Distinguished Young Scholars (61725301, 61925305).
\end{acknowledgements}

%
\section*{Conflict of interest}

The authors declare that they have no conflict of interest.

\bibliographystyle{spmpsci}      
\bibliography{references.bib}   

%
%

\end{document}